\documentclass[letterpaper,journal]{IEEEtran}
\usepackage{cite}
\usepackage{amsmath,amssymb,amsfonts}

\usepackage{amsthm}
\usepackage{algorithmic}
\usepackage{algorithm}
\usepackage{graphicx}
\usepackage{capt-of}
\usepackage{textcomp}
\def\BibTeX{{\rm B\kern-.05em{\sc i\kern-.025em b}\kern-.08em
    T\kern-.1667em\lower.7ex\hbox{E}\kern-.125emX}}
\usepackage{booktabs}
\usepackage{multirow}
\usepackage{stfloats}
\usepackage{flushend}
\usepackage{url}
\usepackage[hidelinks]{hyperref}
\hypersetup{
  pdftitle={Asymmetric Decomposition for Multi-Agent Path Finding with Unassigned Agents},
  pdfauthor={Xi Jin, Chen Qian, Renjun Zhang, Guowei Wu, Xiaobo Li, Jun Jiang, Weiwei Shang},
  pdfkeywords={Conflict-based search, min-cost flow, Multi-Agent Path Finding, multi-robot systems, unassigned agents}
}

\newtheoremstyle{ieeedefinition}%
  {3pt}{3pt}{\normalfont}{}{\itshape}{.}{ }{}
\newtheoremstyle{ieeetheorem}%
  {3pt}{3pt}{\itshape}{}{\itshape}{.}{ }{}

\theoremstyle{ieeedefinition}
\newtheorem{definition}{Definition}
\newtheorem{proposition}{Proposition}
\newtheorem{lemma}{Lemma}
\newtheorem{theorem}{Theorem}
\newtheorem{remark}{Remark}

\newcommand{\sst}{\mathrm{SST}}
\newcommand{\smu}{\mathrm{SMU}}
\newcommand{\alr}{\mathrm{ALR}}
\newcommand{\subfigtext}[1]{{\footnotesize\rmfamily#1}}

\begin{document}
\bstctlcite{IEEEexample:BSTcontrol}

\title{Asymmetric Decomposition for Multi-Agent\\
Path Finding with Unassigned Agents}

\author{%
  \parbox{\textwidth}{\centering
    \fontsize{11}{13}\selectfont
    Xi~Jin\textsuperscript{1,3}, Chen~Qian\textsuperscript{1},
    Renjun~Zhang\textsuperscript{1}, Guowei~Wu\textsuperscript{1},\\
    Xiaobo~Li\textsuperscript{2}, Jun~Jiang\textsuperscript{2,3},
    and Weiwei~Shang\textsuperscript{1,2}\par
    \vspace{0.7em}
    {\fontsize{9}{11}\selectfont
      \textsuperscript{1}University of Science and Technology of China\\
      \textsuperscript{2}State Key Laboratory of Precision and Intelligent Chemistry\\
      \textsuperscript{3}Hefei National Laboratory\par}
    \vspace{0.4em}
    {\fontsize{8}{10}\selectfont\urlstyle{same}
      E-mail: \href{mailto:jinxi@mail.ustc.edu.cn}{\nolinkurl{jinxi@mail.ustc.edu.cn}},
      \href{mailto:wwshang@ustc.edu.cn}{\nolinkurl{wwshang@ustc.edu.cn}}\par}
  }%
}

\maketitle
\suppressfloats[t]

\begin{abstract}
Robots without assigned tasks may obstruct active robots, while routing them to designated parking locations can cause unnecessary travel.
This article proposes flow-oracle conflict-based search (FOCBS) for Multi-Agent Path Finding with unassigned agents (MAPF-UA). FOCBS searches over assigned-agent paths using conflict-based search (CBS) and routes targetless unassigned agents through a min-cost flow oracle.
When the selected assigned-agent paths leave unassigned agents no collision-free paths, flow-conflict branching converts min-cut blockers into assigned-agent constraints, which guide the search without excluding optimal solutions. A batched zero-cost warm start accelerates the oracle, and start-avoiding tie-breaking reduces unassigned-agent movement.
For feasible instances on finite graphs, we prove termination and global minimization of assigned-agent sum of service time, measured by last-arrival times. Unassigned-agent movement is minimized conditional on the selected assigned-agent paths, without a global lexicographic optimality guarantee. Benchmark experiments show that FOCBS scales well with the number of unassigned agents, and ablation studies show that, compared with discarding non-completable assigned-agent paths, flow-conflict branching solves substantially more instances while retaining optimality. Simulation and physical-robot experiments demonstrate plan executability, with laboratory tests reducing fleet travel by 40.8\% relative to the parking-based coordination baseline. The source code and related resources are available at \url{https://github.com/Chin-hsi/focbs}.
\end{abstract}


\section{Introduction}
\label{sec:intro}

\begin{figure}[t]
  \centering
  \includegraphics[width=0.90\columnwidth]{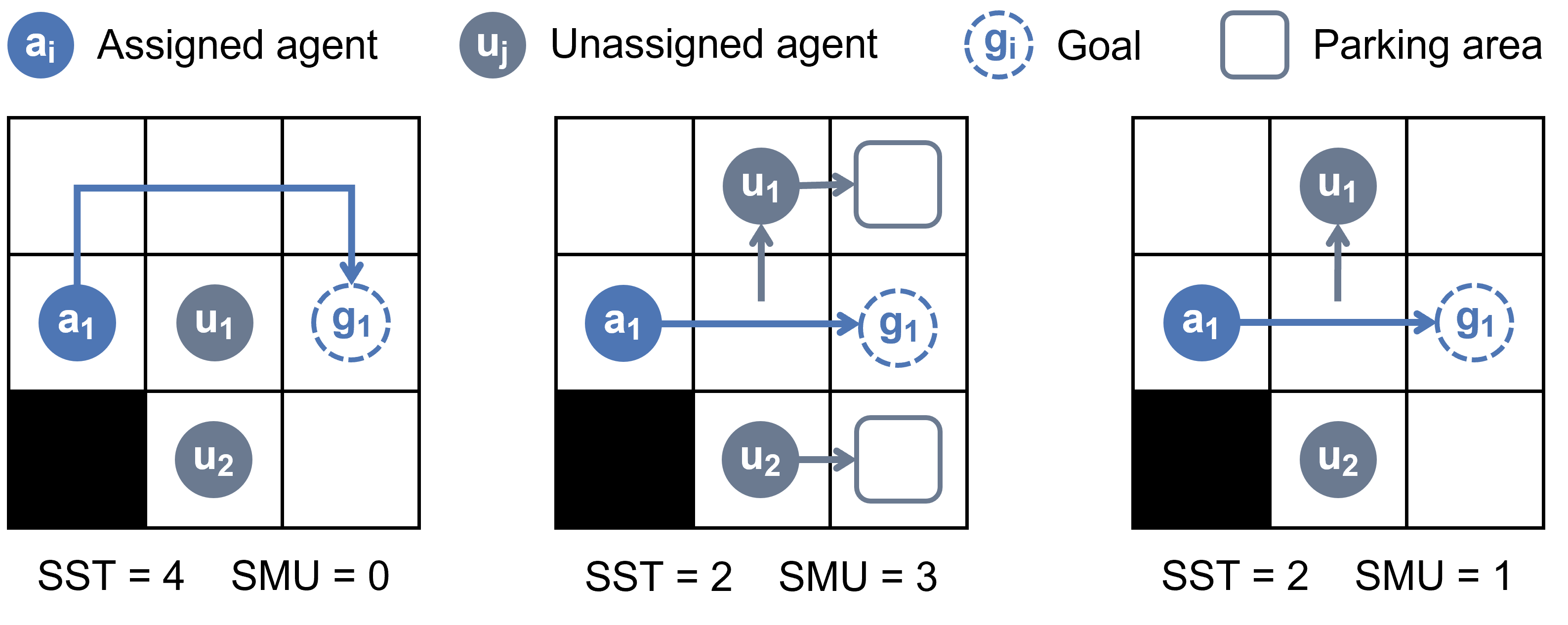}\\[-0.2ex]
  \makebox[0.90\columnwidth][c]{%
    \begin{minipage}[t]{0.2795\columnwidth}\centering
      \makebox[\linewidth][c]{\hspace*{-0.85em}\subfigtext{(a) Static obstacle}}
    \end{minipage}\hfill
    \begin{minipage}[t]{0.2795\columnwidth}\centering
      \makebox[\linewidth][c]{\hspace*{-0.65em}\subfigtext{(b) Parking-area routing}}
    \end{minipage}\hfill
    \begin{minipage}[t]{0.2795\columnwidth}\centering
      \makebox[\linewidth][c]{\hspace*{0.25em}\subfigtext{(c) MAPF-UA}}
    \end{minipage}%
  }
  \caption{A motivating example for MAPF-UA. (a) Treating unassigned agents as static obstacles forces a costly assigned-agent detour. (b) Sending unassigned agents to parking areas clears the path but causes unnecessary movement. (c) MAPF-UA lets only the blocking unassigned agent step aside and stop at any free cell.}
  \label{fig:mapfua_example}
\end{figure}

\IEEEPARstart{M}{ulti-robot} systems improve productivity in automated warehouses, fulfillment centers, manufacturing logistics, and self-driving laboratories~\cite{azadeh2019robotized,lian2021agv,zhou2025multirobot}.
Their operation relies on Multi-Agent Path Finding (MAPF)~\cite{stern2019mapf}, which computes collision-free paths from agents' starts to goals while optimizing a prescribed cost.
For continuously arriving tasks, rolling-horizon methods decompose lifelong warehouse MAPF into successive planning episodes~\cite{li2021lifelong}.
In many deployed mobile robot systems, however, not every robot is assigned a service task in every planning episode.
Some robots must travel from their starts to designated targets to perform transportation, pick-and-place, or laboratory operations, while others have no task targets but still occupy the shared workspace.
Treating these idle robots as static obstacles can block narrow passages, limiting active robots' reachability or forcing long detours; assigning each idle robot a parking location can instead cause unnecessary travel and occupy valuable workspace.
The comparison in Fig.~\ref{fig:mapfua_example} motivates a planner that determines which idle robots need to move, when they should yield, and where they can stop without requiring designated parking locations.

\begin{figure}[t]
  \centering
  \includegraphics[width=\columnwidth]{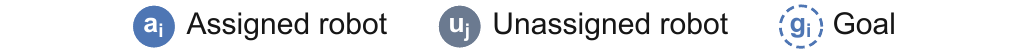}\\[0.4ex]
  \begin{minipage}[t]{0.4914\columnwidth}\centering
    \includegraphics[width=\linewidth]{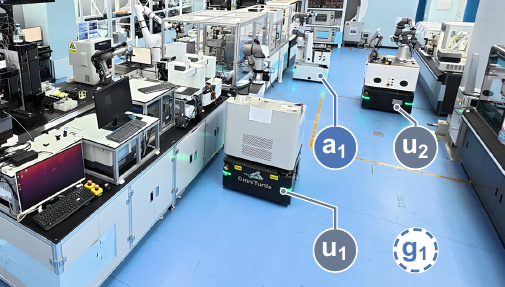}\\[-0.2ex]
    \subfigtext{(a)}
  \end{minipage}\hfill
  \begin{minipage}[t]{0.4914\columnwidth}\centering
    \includegraphics[width=\linewidth]{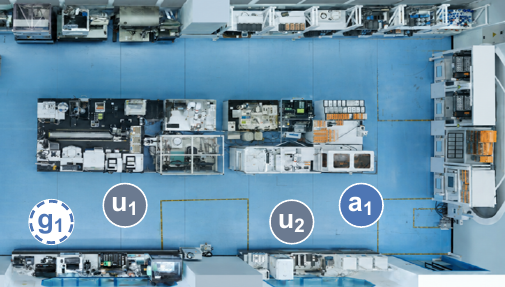}\\[-0.2ex]
    \subfigtext{(b)}
  \end{minipage}\\[0.5ex]
  \includegraphics[width=\columnwidth]{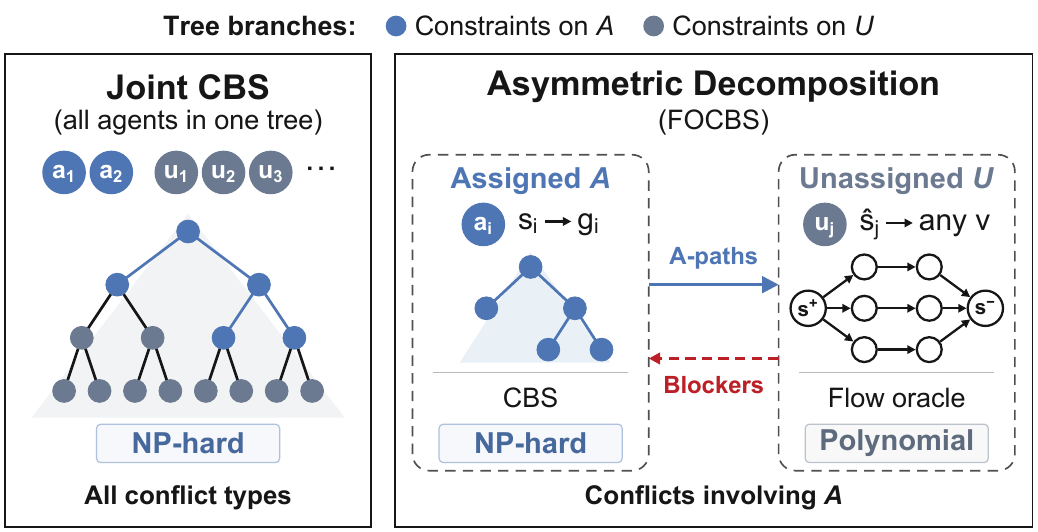}\\[-0.2ex]
  \subfigtext{(c)}
  \caption{MAPF-UA scenario and algorithm comparison. (a) Laboratory scene. (b) Top view. (c) Algorithm comparison: joint CBS searches over all agents in one constraint tree; FOCBS combines CBS over assigned agents with polynomial-time flow routing for unassigned agents under fixed assigned-agent paths.}
  \label{fig:asymmetric_decomposition}
\end{figure}

Multi-Agent Path Finding with unassigned agents (MAPF-UA)~\cite{felner2026mapfua} captures this setting.
Assigned agents have designated targets, whereas unassigned agents have only starts and may terminate at any vertex; all agents must remain collision-free.
Minimizing assigned-agent service time alone may move unassigned agents unnecessarily.
Subject to collision-free paths for all agents, we therefore first minimize the sum of service time (SST) for assigned agents and then minimize the sum of moves by unassigned agents (SMU) conditional on the selected assigned-agent paths.

Because MAPF-UA generalizes MAPF, solving it optimally is NP-hard~\cite{yu2016intractability}.
Unassigned agents add two difficulties.
Their terminal vertices are decision variables, and permuting them over the same vertices yields equivalent states.
Moreover, static unassigned agents may greatly increase service time or render dense instances unsolvable~\cite{felner2026mapfua}.
In dense instances, scarce free vertices leave unassigned agents little space to yield.

Conflict-based search (CBS)~\cite{sharon2015cbs} uses a two-level search.
The high level resolves conflicts by creating child nodes in a constraint tree, each imposing a constraint on one of the conflicting agents; the low level replans that agent's path under the updated constraints.
Applying CBS jointly to both agent types places assigned agents and unassigned agents in the same high-level constraint tree (Fig.~\ref{fig:asymmetric_decomposition}(c), left).
This direct application does not fully exploit the targetless, interchangeable nature of unassigned agents.
As unassigned-agent density grows, branching on conflicts involving them substantially increases high-level search overhead, creating a scalability bottleneck.

Optimal CBS extensions reduce search effort via stronger reasoning and splitting, including improved CBS (ICBS)~\cite{boyarski2015icbs}, disjoint splitting~\cite{li2019disjoint}, dependency-graph heuristics~\cite{li2019improved}, and symmetry reasoning~\cite{li2021symmetry}.
In-depth cluster reasoning strengthens CBS heuristics by identifying conflicts involving more than two agents~\cite{lee2026indepth}.
Enhanced CBS (ECBS)~\cite{barer2014ecbs} and explicit estimation CBS (EECBS)~\cite{li2021eecbs} improve efficiency with bounded-suboptimality guarantees.
For multi-agent combinatorial path finding, CBSS-A and CBSS-AF combine approximate target sequencing with collision-free routing~\cite{ren2025boundedmcpf}.
The cited formulations consider agents with prescribed targets.

\suppressfloats[t]

Other methods emphasize scalable coordination or feasible-path search.
Priority inheritance with backtracking (PIBT)~\cite{okumura2022pibt} provides fast online coordination, the LaCAM family~\cite{okumura2024engineeringlacam} uses lazy constraint addition, and the multi-agent corridor generating algorithm (MACGA)~\cite{pertzovsky2025cga} constructs target-directed corridors and evacuates blocking agents.
The acceptance-rejection sampling-guided discrete rapidly-exploring random tree (adRRT)~\cite{xin2026adrrt} searches for feasible paths through randomized expansion.
Learning-based methods, including pathfinding via reinforcement and imitation multi-agent learning (PRIMAL)~\cite{sartoretti2019primal} and transformer-based approaches~\cite{chen2023transformer}, train decentralized policies for online decisions.
Chen et al.~\cite{chen2023hscl} combine deep reinforcement learning with hot supervision contrastive loss to learn coordination without interagent communication.
These methods provide efficient coordination mechanisms, but their original formulations do not explicitly minimize unassigned-agent movement.

Network-flow and decomposition methods offer another way to organize multi-agent planning.
Time-expanded networks (TENs) encode routing as network flow~\cite{yu2013network}, and conflict-based min-cost flow (CBM)~\cite{ma2016cbm} combines CBS between teams with min-cost flow within teams having prescribed target sets.
For labeled robots, multicommodity-flow encodings support integer-programming algorithms for several path-cost objectives~\cite{yu2016optimal}.
Branch-and-cut-and-price also uses decomposition~\cite{lam2022bcp}, while anonymous MAPF exploits agent interchangeability with respect to goals~\cite{ali2024amapf}.
Layered decomposition for large agents partitions the planning problem and merges subproblems when needed to restore solvability~\cite{yao2026decomposing}.
Lexicographic bi-objective MAPF~\cite{zhao2025lcbs} and multi-objective CBS~\cite{ren2023bbmocbs} address ordered objectives and Pareto-optimal solution sets, respectively.
These formulations assign each agent a target or each group a target set, whereas unassigned agents may stop at any free vertex.

MAPF-UA is closely related to MAPF with the reachability objective (MAPF-RO)~\cite{gabay2026reachability}, in which each agent must visit its target at least once but may subsequently leave it.
Gabay et al.~\cite{gabay2026reachability} reduce the reachability-based MAPF-UA formulation to MAPF-RO by assigning each unassigned agent its start vertex as its target; this reduction preserves soundness, optimality, and completeness under their first-arrival-based SST objective.
The resulting MAPF-RO instance treats both groups as agents with targets, and the originally unassigned agents satisfy their targets at the initial time.
Their subsequent movement therefore contributes nothing to the standard MAPF-RO objective, which neither penalizes that movement nor optimizes $\smu$.

In MAPF-UA, unassigned agents are commonly treated as static obstacles, assigned surrogate targets such as their starts or parking locations, or planned individually within a joint search.
These treatments do not fully exploit the targetless, interchangeable nature of unassigned agents.
Decomposition is a natural approach: with assigned-agent paths fixed, targetless unassigned agents can be routed by single-commodity min-cost flow in polynomial time.
However, $\sst$-optimal assigned-agent paths computed without unassigned agents may occupy the vertices and time steps they need to yield, leaving no collision-free paths for unassigned agents, and the failure alone does not identify which paths to revise.
The central challenge is to exploit the targetless structure for scalability while using such failures to guide the assigned-agent search.

Feeding subproblem infeasibility back to a higher-level search follows the decomposition principle of Benders~\cite{benders1962partitioning}.
Logic-based Benders decomposition derives constraints from subproblem inference~\cite{hooker2003logic}, and combinatorial Benders cuts exclude combinations of decisions that make a subproblem infeasible~\cite{codato2006combinatorial}.
FOCBS adopts a similar feedback idea for MAPF-UA: a minimum cut~\cite{ford1956maximal} identifies assigned-agent reservations whose joint retention prevents unassigned-agent routing, and their negations form a CBS branching disjunction.
Standard CBS splitting enforces such a disjunction for a pairwise conflict~\cite{sharon2015cbs}; for MAPF-UA, the key step is to show that the flow-derived disjunction covers every feasible completion under the current constraints, including those with longer horizons.

Our contributions are as follows.
\begin{enumerate}
  \item We formulate sequential MAPF-UA and establish its asymmetric decomposition: finding $\sst$-optimal completable assigned-agent paths is NP-hard, whereas unassigned-agent routing under fixed assigned-agent paths is solvable by min-cost flow in time polynomial in TEN size.
  \item We propose flow-oracle conflict-based search (FOCBS), using CBS for assigned agents and flow-conflict branching to convert min-cut blockers into a CBS disjunction. We prove that this disjunction preserves every feasible completion, including those with longer horizons, which yields termination and sequential optimality for feasible instances on finite graphs.
  \item We develop a batched zero-cost warm start to accelerate the flow oracle, and a start-avoiding tie-breaking rule to reduce $\smu$ without compromising $\sst$ optimality.
  \item We evaluate FOCBS on standard benchmarks and validate path executability in warehouse simulation and physical-robot experiments.
\end{enumerate}

The remainder of this article is organized as follows.
Section~\ref{sec:problem} formulates MAPF-UA and the asymmetric decomposition.
Section~\ref{sec:algorithm} presents the FOCBS algorithm and the flow oracle.
Section~\ref{sec:experiments} evaluates FOCBS on standard benchmarks, reports ablation experiments and runtime analysis, and validates path executability through simulation and real-world experiments.
Section~\ref{sec:conclusion} concludes this article.

\section{Problem Formulation}
\label{sec:problem}

We formulate sequential MAPF-UA, which minimizes $\sst$ globally and then $\smu$ for the selected assigned-agent paths. Finding $\sst$-optimal completable assigned-agent paths is NP-hard, whereas unassigned-agent routing under fixed ones is polynomial-time solvable by min-cost flow.

\begin{definition}[MAPF-UA]
A MAPF-UA instance consists of an undirected graph $G=(V,E)$ and disjoint sets of assigned agents $A=\{a_1,\ldots,a_n\}$ and unassigned agents $U=\{u_1,\ldots,u_m\}$.
Each assigned agent $a_i$ has a start $s_i\in V$ and target $g_i\in V$; each unassigned agent $u_j$ has a start $\hat{s}_j\in V$ but no target.
All start vertices are distinct.
A path is a vertex sequence, indexed by time step, whose consecutive vertices are either identical (a wait) or adjacent in $G$ (a move).
We call assigned-agent paths \emph{A-paths} and unassigned-agent paths \emph{U-paths}.
A complete plan consists of an A-path set $\pi^A=(\pi_i^A)_{a_i\in A}$ and a U-path set $\pi^U=(\pi_j^U)_{u_j\in U}$, where $\pi_i^A$ starts at $s_i$ and $\pi_j^U$ starts at $\hat{s}_j$.
A vertex conflict occurs when two agents occupy the same vertex at the same time step; an edge conflict occurs when two agents traverse the same edge in opposite directions between the same two consecutive time steps.
A complete plan is feasible if it is conflict-free, each A-path reaches its target and stays there from some finite time onward, and each U-path eventually stays at some vertex.
\end{definition}

The agent-location ratio (ALR) is defined as $\alr=(n+m)/|V|$~\cite{felner2026mapfua}.
As $\alr$ approaches one, free vertices become scarce, and unassigned agents more often need to yield, which makes unassigned-agent routing harder.

The last-arrival time is the earliest time step from which $a_i$ stays at $g_i$ forever:
\begin{equation}
  t_i^{\mathrm{last}}(\pi_i^A):=\min\{t\in\mathbb Z_{\ge0}:\pi_i^A(\tau)=g_i,\ \forall\tau\in\mathbb Z_{\ge t}\}.
  \label{eq:last_arrival}
\end{equation}
SST uses this last-arrival time because an assigned agent must remain at its target to perform its task; the reachability formulation of~\cite{gabay2026reachability} instead uses first-arrival times and permits subsequent departure.
The A-paths determine the horizon
\begin{equation}
  T(\pi^A)=\max_{a_i\in A} t_i^{\mathrm{last}}(\pi_i^A), \label{eq:node_horizon}
\end{equation}
after which every assigned agent waits at its target.

For a complete plan $\pi=(\pi^A,\pi^U)$, the two costs are
\begin{align}
  \sst(\pi) &= \textstyle\sum_{a_i\in A} t_i^{\mathrm{last}}(\pi_i^A), \label{eq:sst} \\
  \smu(\pi) &= \textstyle\sum_{u_j\in U}\sum_{t=1}^{T(\pi^A)}
  \mathbf{1}\{\pi_j^U(t)\neq\pi_j^U(t{-}1)\}. \label{eq:smu}
\end{align}
Here $\sst$ is the sum of service time for assigned agents~\cite{felner2026mapfua}, also written $\sst(\pi^A)$, and $\smu$ is the sum of moves by unassigned agents through $T(\pi^A)$.
In Fig.~\ref{fig:mapfua_example}, the static-obstacle, parking, and MAPF-UA plans have $(\sst,\smu)=(4,0)$, $(2,3)$, and $(2,1)$, respectively.
The last two plans share the same A-path, but parking moves unassigned agents beyond what yielding requires.
Let $\mathcal A$ be the set of mutually conflict-free A-path sets in which each $a_i$ starts at $s_i$, reaches $g_i$, and waits there from its last arrival through $T(\pi^A)$.
For $\pi^A\in\mathcal A$, let $\mathcal R(\pi^A)$ be the set of U-path sets that, over time steps $0,\ldots,T(\pi^A)$, are conflict-free among themselves and with $\pi^A$. Define
\begin{equation}
  \mathcal A_{\mathrm{comp}}
  =\{\pi^A\in\mathcal A:\mathcal R(\pi^A)\neq\emptyset\}.
  \label{eq:acomp}
\end{equation}
We call $\pi^A$ completable if $\mathcal R(\pi^A)\neq\emptyset$, and each $\pi^U\in\mathcal R(\pi^A)$ a completion of $\pi^A$.
After $T(\pi^A)$, all assigned agents wait at their targets; thus, a conflict-free configuration at $T(\pi^A)$ remains conflict-free if all agents wait thereafter.
Hence, an instance is feasible if and only if $\mathcal A_{\mathrm{comp}}\neq\emptyset$, and restricting U-paths to $0,\ldots,T(\pi^A)$ loses no feasible plan.
For a feasible instance, let
\begin{align}
  C^* &= \min_{\pi^A\in\mathcal A_{\mathrm{comp}}}\sst(\pi^A), \label{eq:cstar} \\
  \mathcal A^* &= \arg\min_{\pi^A\in\mathcal A_{\mathrm{comp}}}\sst(\pi^A). \label{eq:astar}
\end{align}
For each $\pi^A\in\mathcal A_{\mathrm{comp}}$, define
\begin{align}
  Q(\pi^A) &= \min_{\pi^U\in\mathcal R(\pi^A)}\smu((\pi^A,\pi^U)), \label{eq:qvalue} \\
  \mathcal R^*(\pi^A) &= \arg\min_{\pi^U\in\mathcal R(\pi^A)}\smu((\pi^A,\pi^U)). \label{eq:rstar}
\end{align}

\begin{definition}[Sequential MAPF-UA]
\label{def:seq_mapfua}
A complete plan $\hat\pi=(\hat\pi^A,\hat\pi^U)$ is sequentially optimal if
\begin{equation}
  \hat\pi^A\in\mathcal A^*,
  \qquad
  \hat\pi^U\in\mathcal R^*(\hat\pi^A).
  \label{eq:seq_optimality}
\end{equation}
Equivalently, $\sst(\hat\pi)=C^*$ and $\smu(\hat\pi)=Q(\hat\pi^A)$.
\end{definition}

For comparison, global lexicographic optimization first minimizes $\sst$ and then minimizes $\smu$ over all $\sst$-optimal feasible plans.
A globally lexicographically optimal plan $\hat\pi$ therefore satisfies
\begin{equation}
  \sst(\hat\pi)=C^*,
  \qquad
  \smu(\hat\pi)=\min_{\pi^A\in\mathcal A^*}Q(\pi^A).
  \label{eq:lex_optimality}
\end{equation}

Both objectives attain the same globally minimum $\sst$.
Sequential optimality minimizes $\smu$ conditional on one selected $\hat\pi^A\in\mathcal A^*$, whereas global lexicographic optimality also minimizes $Q$ over $\mathcal A^*$.
This additional optimization can increase the search effort.

We adopt the sequential objective because completing assigned tasks is the primary concern in the target applications, while unassigned-agent movement is secondary.
A small amount of extra movement by agents without current tasks is often acceptable in exchange for improved scalability, provided that $\sst$ remains globally optimal.
This tradeoff is evaluated in Section~\ref{sec:exp_composition_revised}.

\begin{remark}[Completability coupling]
\label{rem:coupling}
Let $C_{\mathcal A}=\min_{\pi^A\in\mathcal A}\sst(\pi^A)$.
Since $\mathcal A_{\mathrm{comp}}\subseteq\mathcal A$, we have $C_{\mathcal A}\le C^*$.
However, $\sst$-optimal A-paths in $\mathcal A$ may occupy the vertices and time steps that unassigned agents need to yield, so that $\mathcal R(\pi^A)=\emptyset$.
In Fig.~\ref{fig:flow_conflict}, conflict-free A-paths with $\sst=C_{\mathcal A}=4$ admit no U-paths, whereas $C^*=5$.
A two-stage method that fixes such A-paths fails; the search over A-paths must therefore account for completability.
\end{remark}

\begin{proposition}[Asymmetric structure]
\label{prop:decomposition}
This formulation has the following structural properties:
(a) finding an A-path set in $\mathcal A^*$ is NP-hard; and
(b) for fixed $\pi^A\in\mathcal A$, finding U-paths in $\mathcal R^*(\pi^A)$, or reporting $\mathcal R(\pi^A)=\emptyset$, takes time polynomial in $|V|$, $|E|$, and $T(\pi^A)$.
\end{proposition}
\begin{proof}
For part~(a), when $m=0$, every A-path set in $\mathcal A$ is completable; hence, $\mathcal A^*$ is the set of sum-of-costs-optimal MAPF solutions, whose computation is NP-hard~\cite{yu2016intractability}.
Thus, finding an A-path set in $\mathcal A^*$, and hence solving sequential MAPF-UA, is NP-hard.
For part~(b), fixed A-paths determine which vertices and edge traversals are occupied at each time step, while the targetless unassigned agents form a single commodity.
By the construction in Section~\ref{sec:ten}, every U-path set in $\mathcal R(\pi^A)$ induces an integral flow of value $m$ with cost equal to its $\smu$, and every integral flow of value $m$ can be converted into a U-path set in $\mathcal R(\pi^A)$ without increasing cost.
Thus, $\mathcal R(\pi^A)\neq\emptyset$ if and only if a flow of value $m$ exists; when such a flow exists, its minimum cost equals $Q(\pi^A)$.
The TEN has $O((|V|+|E|)(T(\pi^A)+1))$ vertices and arcs. Min-cost flow on this network can be solved in polynomial time~\cite{ahuja1993network}.
\end{proof}

\section{FOCBS Algorithm}
\label{sec:algorithm}

\begin{figure}[!t]
  \centering
  \includegraphics[width=\columnwidth]{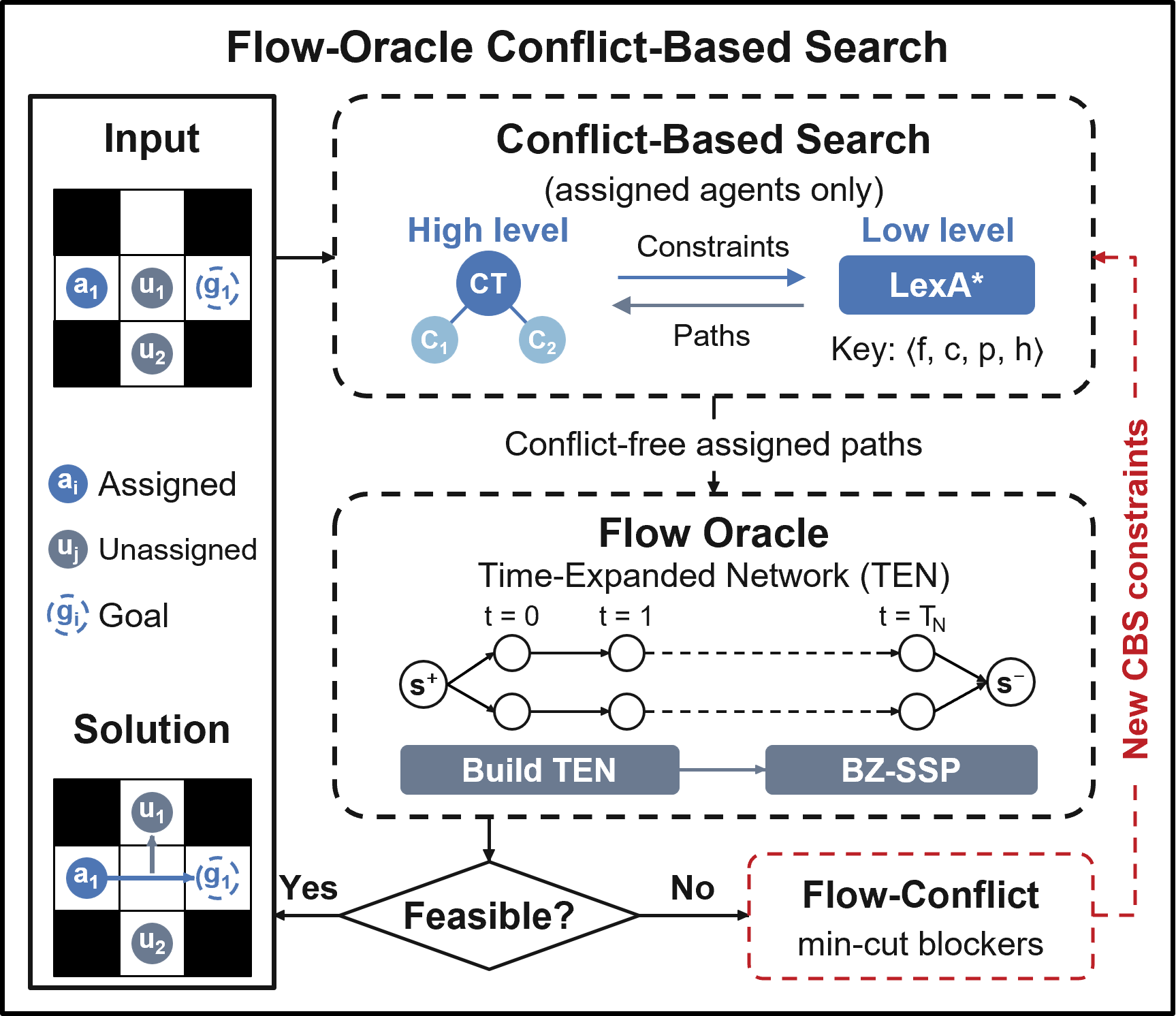}
  \caption{Overview of FOCBS. CBS constrains only assigned agents. For each conflict-free A-path set, the flow oracle seeks conflict-free U-paths. If none exist, it returns min-cut blockers to generate new CBS constraints.}
  \label{fig:overview}
\end{figure}

FOCBS solves sequential MAPF-UA with a CBS search over assigned agents and a flow oracle for unassigned-agent routing (Fig.~\ref{fig:overview}).
Following Proposition~\ref{prop:decomposition}, it exploits the absence of individual targets: a single-commodity flow decides the terminal vertices of unassigned agents implicitly; hence, the high-level search constrains only A-paths.
Each constraint-tree (CT) node $N$ stores only constraints and A-paths $\pi^A_N$, whose horizon is $T_N:=T(\pi^A_N)$; the open list OPEN is ordered by $\sst$.
FOCBS first resolves conflicts among assigned agents with standard CBS splits.
Once the A-paths of a node are mutually conflict-free, the oracle treats them as reservations and computes minimum-$\smu$ U-paths in a time-expanded network (TEN).
If such U-paths exist, FOCBS returns the complete plan; otherwise, flow-conflict branching converts min-cut blockers into constraints on assigned agents and replans their paths.
The two groups thus remain coupled: unassigned agents determine the flow value, the TEN size, and whether the selected A-paths are completable.

\subsection{Assigned-Agent CBS with Flow Oracle}
\label{sec:cbs_oracle}

Algorithm~\ref{alg:folcbs} presents FOCBS: assigned-agent CBS~\cite{sharon2015cbs} forms the search backbone, while lines~\ref{line:focbs_oracle}--\ref{line:focbs_flow_branch} implement the flow-oracle extension.
If FOCBS returns \textsc{InstanceInfeasible}, the MAPF-UA instance has no solution. For feasible instances, FOCBS terminates and returns a solution (Theorem~\ref{thm:optimality}); termination is not guaranteed for infeasible instances.

FOCBS applies start-avoiding tie-breaking only among cost-equivalent choices, preferring A-paths or CT nodes with fewer visits to unassigned-agent starts.
When an A-path visits an unassigned agent's start, that agent must leave it at least once; fewer such visits therefore tend to reduce $\smu$.
Let $\hat{S}=\{\hat{s}_j:u_j\in U\}$ be the unassigned-agent start set.
At the low level, FOCBS computes A-paths with lexicographic A* (LexA*) using the key $\langle f,c,p,h\rangle$ and a conflict-avoidance table (CAT)~\cite{sharon2015cbs}. Here, $f{=}g{+}h$, $c$ counts CAT conflicts, $p$ counts time-indexed visits to $\hat{S}$ along the partial path, and $h$ is the target-distance heuristic.
Because $p$ is ordered after $f$ and $c$, it only breaks ties among partial paths with equal $f$ and $c$.
At the high level, CT nodes with equal $\sst$ are ranked by $(d,q)$: $d$ is the number of distinct conflicting assigned-agent pairs, and $q$ counts visits to unassigned-agent starts up to and including each assigned agent's last arrival:
\[
q=\sum_{a_i\in A}\sum_{t=0}^{t_i^{\mathrm{last}}(\pi_{N,i}^A)}
\mathbf{1}\{\pi_{N,i}^A(t)\in\hat{S}\}.
\]
Neither tie-breaking key alters the $\sst$-optimality guarantee.
Each low-level invocation returns exactly one path of minimum last-arrival time under the current constraints, with LexA* selecting among equal-cost alternatives according to the stated low-level ordering. The remaining alternatives are not materialized as sibling CT nodes: CT children are created only by adding a CBS or flow-conflict constraint, after which replanning may select a different equal-cost path.

\begin{algorithm}[t]
\caption{FOCBS.}
\label{alg:folcbs}
\begin{algorithmic}[1]
\STATE \textbf{Input:} $G$, assigned agents $(A,s,g)$, starts $\hat S=\{\hat{s}_j\}_{j=1}^m$
\STATE \textbf{Output:} solution $(\pi^A,\pi^U)$, or \textsc{InstanceInfeasible}
\STATE Initialize unconstrained $N_0$ with shortest paths for $A$
\STATE \textbf{if} any root path is absent \textbf{then return} \mbox{\textsc{InstanceInfeasible}}
\STATE $\mathit{OPEN} \leftarrow \{N_0\}$, ordered by $\sst$
\WHILE{$\mathit{OPEN} \neq \emptyset$}
  \STATE $N \leftarrow \mathit{OPEN}.\textsc{Pop}()$
  \IF{$\textsc{FindConflict}(N.\mathit{paths})$ returns conflict $\chi$}
    \STATE $\mathcal C \leftarrow \textsc{CBSConstraints}(\chi)$
  \ELSE
    \STATE Run $\textsc{FlowOracle}(G,N.\mathit{paths},T_N,\hat S)$\label{line:focbs_oracle}
    \STATE \textbf{if} $\pi^U$ is returned \textbf{then return} $(N.\mathit{paths},\pi^U)$
    \STATE $\mathcal C \leftarrow \textsc{BlockerConstraints}(\text{returned blockers})$\label{line:focbs_flow_branch}
  \ENDIF
  \FOR{each constraint $c \in \mathcal C$}
    \STATE Copy $N$ to $N'$, add $c$, and replan agent $a(c)$
    \IF{replanning succeeds}
      \STATE Update $N'.\mathit{paths}$, $\sst$, and tie-breaking keys
      \STATE Insert $N'$ into $\mathit{OPEN}$
    \ENDIF
  \ENDFOR
\ENDWHILE
\RETURN \textsc{InstanceInfeasible}
\end{algorithmic}
\end{algorithm}

\subsection{Time-Expanded Network Construction}
\label{sec:ten}

\begin{figure}[t]
  \centering
  \includegraphics[width=\columnwidth]{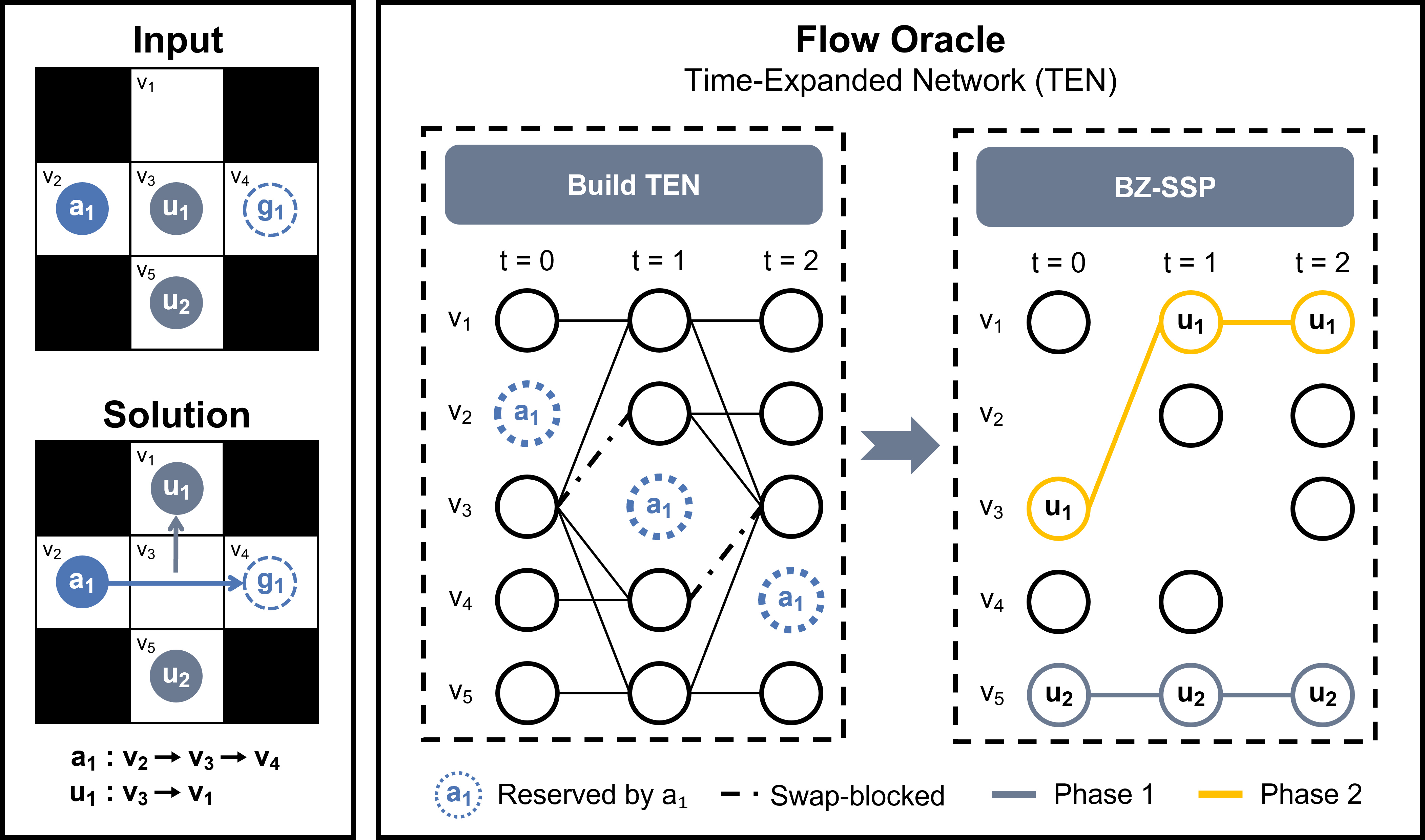}\\[-0.2ex]
  \makebox[\columnwidth][c]{%
    \begin{minipage}[t]{0.2506\columnwidth}\centering
      \subfigtext{(a) Problem}
    \end{minipage}\hfill
    \begin{minipage}[t]{0.6906\columnwidth}\centering
      \subfigtext{(b) Flow solving}
    \end{minipage}%
  }
  \caption{TEN for fixed A-paths. (a) Instance and U-paths. (b) TEN with reserved nodes and opposite-move arcs removed. Phase~1 inserts zero-cost waiting paths at unreserved starts; Phase~2 augments the remaining flow. Vertex-splitting and merge--split gadget nodes are omitted for clarity.}
  \label{fig:flow_oracle_ten}
\end{figure}

For fixed A-paths $\pi^A_N$, the oracle uses a TEN~\cite{yu2013network,ma2016cbm} with layers $t=0,\ldots,T_N$ and arcs between consecutive layers.
Fig.~\ref{fig:flow_oracle_ten} illustrates the construction. The part of the TEN that does not depend on the A-paths, called the skeleton, contains:

\begin{itemize}
  \item For each vertex-time pair $(v,t)$, two nodes $v^{\mathrm{in}}_t,v^{\mathrm{out}}_t$ and a unit-capacity, zero-cost arc $v^{\mathrm{in}}_t\to v^{\mathrm{out}}_t$ enforcing \mbox{vertex capacity}.
  \item \emph{Wait arcs}: $v^{\mathrm{out}}_t \to v^{\mathrm{in}}_{t+1}$, capacity~1, cost~0.
  \item \emph{Merge--split edge gadgets}: For each edge $e=\{v,w\}\in E$ and interval $[t,t{+}1]$, two auxiliary nodes $e_t^{\mathrm{in}},e_t^{\mathrm{out}}$ and five unit-capacity arcs~\cite{yu2013network,ma2016cbm}:
  \[
  \begin{gathered}
    v_t^{\mathrm{out}}\to e_t^{\mathrm{in}},\quad
    w_t^{\mathrm{out}}\to e_t^{\mathrm{in}},\quad
    e_t^{\mathrm{in}}\to e_t^{\mathrm{out}},\\
    e_t^{\mathrm{out}}\to v_{t+1}^{\mathrm{in}},\quad
    e_t^{\mathrm{out}}\to w_{t+1}^{\mathrm{in}}.
  \end{gathered}
  \]
  For $\smu$, the shared central arc $e_t^{\mathrm{in}}\to e_t^{\mathrm{out}}$ has cost~1 and the four attachment arcs cost~0, as in the distance-minimizing formulation of~\cite{yu2013network}.
  Moves $v\to w$ and $w\to v$ traverse the gadget between opposite endpoints and share its central capacity-1 arc, preventing simultaneous opposite moves.
  Every move passes through a gadget.
  \item Source arcs $s^+\to (\hat{s}_j)^{\mathrm{in}}_0$ for each unassigned-agent start and sink arcs $v_{T_N}^{\mathrm{out}}\to s^-$ for every $v\in V$, all with unit capacity and zero cost. The sink arcs allow unassigned agents to choose their terminal vertices.
\end{itemize}

Each A-path reserves its vertex at every time step, including its waits at the target through $T_N$.
The ordinary TEN removes the nodes of every reserved vertex-time pair $(v,t)$ and their incident arcs.
If $a_i$ traverses $v\to w$ during $[t,t{+}1]$, the ordinary TEN also removes $w_t^{\mathrm{out}}\to e_t^{\mathrm{in}}$, which would allow the opposite move $w\to v$.
The diagnostic TEN $\mathcal N^{\mathrm{diag}}$ keeps the whole skeleton but sets the capacity of each reserved vertex arc $v^{\mathrm{in}}_t\to v^{\mathrm{out}}_t$ and each such opposite-move arc to zero, labeling them by $(a_i,v,t)$ and $(a_i,v,w,t)$, respectively.
Because reserved vertex-time pairs are removed or have zero capacity, a vertex reserved at $T_N$ cannot serve as a terminal vertex.
For a fixed horizon, different A-paths change only arc capacities; therefore, the capacity of the same cut can be compared across A-paths.

The construction gives a two-way correspondence between paths and flow.
Sending one unit from each unassigned-agent start along its vertex, wait, and gadget arcs maps every U-path set in $\mathcal R(\pi^A_N)$ to a value-$m$ integral flow of equal $\smu$ cost.
Vertex-capacity and central gadget arcs exclude vertex and edge conflicts, respectively, while reservations exclude conflicts with $\pi^A_N$.

Conversely, the acyclic TEN allows every value-$m$ integral flow to decompose into $m$ unit $s^+$--$s^-$ paths, which project onto $G$ after deleting split and gadget nodes.
Call an integral flow canonical if every gadget traversal exits at the endpoint opposite to its entry.
The merge--split gadget also admits a same-endpoint traversal, or turnback, which projects to a wait.
If a flow contains a turnback at $v$, the direct wait arc $v_t^{\mathrm{out}}\to v_{t+1}^{\mathrm{in}}$ is unused; otherwise the unit vertex capacity at $(v,t)$ would be exceeded.
Replacing the turnback by this wait arc preserves feasibility and reduces its cost from~1 to~0.
Repeating this replacement yields a canonical flow, whose cost equals the $\smu$ of its projected paths.
Every minimum-cost integral flow is therefore canonical, and the minimum flow cost equals the minimum $\smu$.
Because unassigned agents may stop at any unreserved vertex, the horizon $T_N$ suffices for fixed A-paths: any U-path set over $0,\ldots,T_N$ can be extended by zero-cost waiting, and any longer conflict-free U-path set can be restricted to $0,\ldots,T_N$ without increasing $\smu$.

The oracle uses successive shortest path (SSP)~\cite{ahuja1993network} on a TEN of $O(L_N)$ vertices and arcs, where $L_N=(|V|+|E|)(T_N+1)$.
One oracle call takes
$O(L_N+mL_N\log L_N)$ time using at most $m$ reduced-cost Dijkstra
searches with binary heaps.

\subsection{Flow-Conflict Branching}
\label{sec:flow_conflict}

\begin{figure}[t]
  \centering
  \includegraphics[width=\columnwidth]{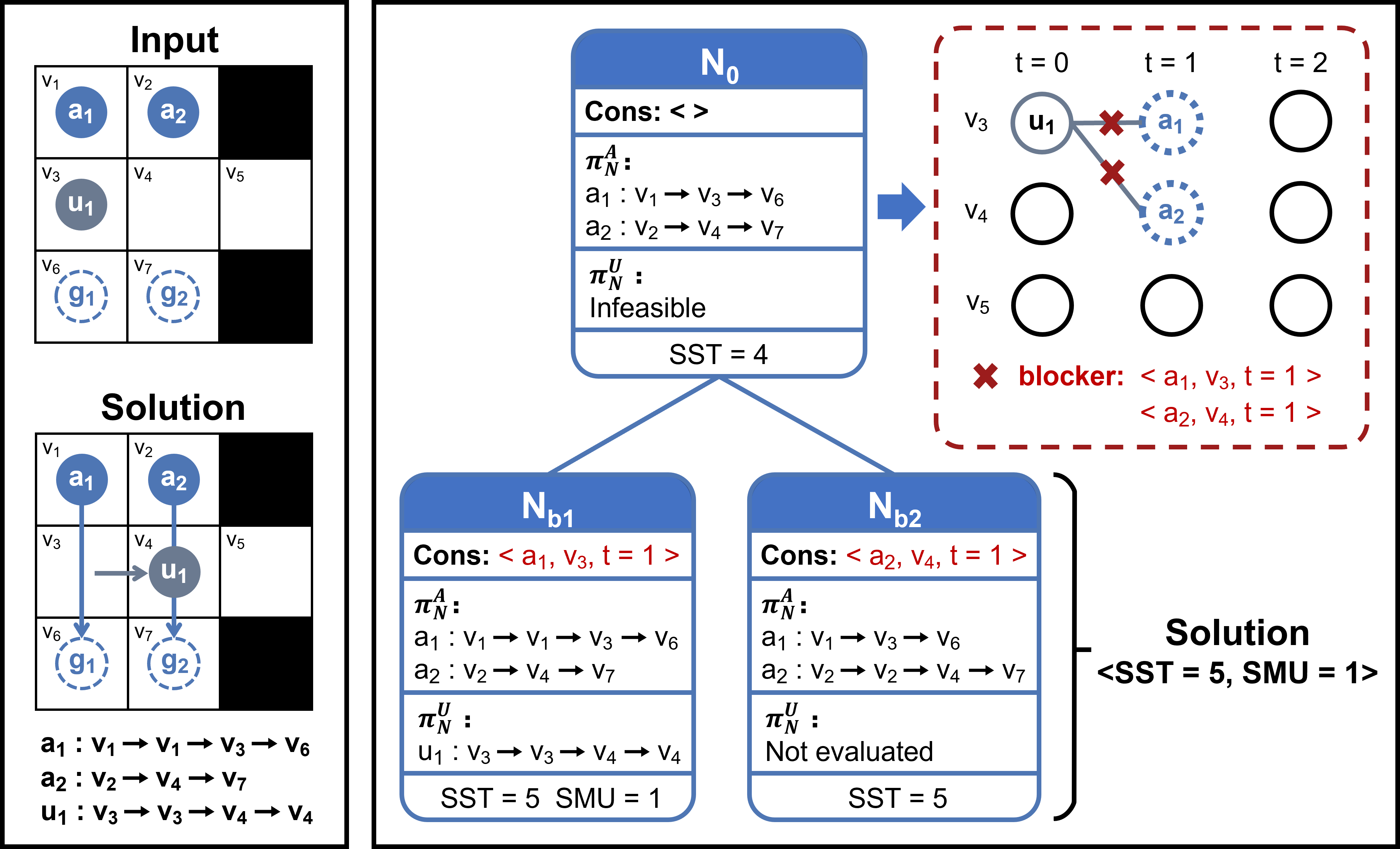}\\[-0.2ex]
  \makebox[\columnwidth][c]{%
    \begin{minipage}[t]{0.2506\columnwidth}\centering
      \subfigtext{(a) Problem}
    \end{minipage}\hfill
    \begin{minipage}[t]{0.7306\columnwidth}\centering
      \subfigtext{(b) Flow-conflict branching}
    \end{minipage}%
  }
  \caption{Partial illustration of flow-conflict branching. (a) Instance and final feasible paths. (b) Root A-paths have $\sst=4$; ``infeasible'' means that no conflict-free U-paths exist under these A-paths. Only the two min-cut vertex blockers at $t=1$ ($a_1$ at $v_3$ and $a_2$ at $v_4$) and their CBS children are shown; FOCBS branches on every branchable blocker in the cut. In each CT node $N$, $\pi^A_N$ denotes its stored A-paths and $\pi^U_N$ the U-paths returned by the oracle.}
  \label{fig:flow_conflict}
\end{figure}

As Remark~\ref{rem:coupling} shows, mutually conflict-free A-paths may leave no U-paths.
An oracle failure rules out only the current A-paths, not every A-path set satisfying the node's constraints; stopping the search or pruning the node may discard all feasible plans below it.
FOCBS therefore extracts from the failure the assigned-agent reservations that must change.

FOCBS branches when assigned agents conflict with each other or when the oracle cannot find conflict-free paths for all unassigned agents.
$\textsc{FlowOracle}$ constructs the ordinary TEN from its inputs and invokes BZ-SSP (Algorithm~\ref{alg:bzssp}) for routing.
If BZ-SSP cannot find conflict-free paths for all unassigned agents, $\textsc{FlowOracle}$ builds the diagnostic TEN and computes a maximum flow on it.
The source-reachable set in the residual network of this maximum flow defines a minimum $s^+$--$s^-$ cut~\cite{ahuja1993network}; every labeled zero-capacity arc in this cut is a blocker.
$\textsc{FlowOracle}$ returns the blockers from the min-cut. Algorithm~\ref{alg:folcbs} converts these blockers into new CBS constraints through $\textsc{BlockerConstraints}$.

A blocker is branchable unless it is a $t{=}0$ vertex blocker, which the starts fix, or a reservation that the current branch already forbids.
For each branchable blocker, FOCBS creates a CBS child that forbids it: a vertex blocker yields the constraint $(a_i,v,t)$, and an edge blocker forbids the traversal $(a_i,v,w,t)$.
By Lemma~\ref{lem:blocker}, every feasible plan under the current constraints releases at least one of these blockers; hence, the children together cover all such plans.
Releasing one blocker need not restore feasibility, since another reservation or cut may remain; each child is therefore processed by the same CBS--oracle procedure.
If no branchable blocker remains, the same lemma shows that no feasible plan satisfies the node's constraints, and the node is discarded.

To express this feedback explicitly, let $\mathcal B$ be the branchable blocker set and $x_b\in\{0,1\}$ indicate whether a candidate A-path set retains the agent-specific vertex occupancy or edge traversal labeled by $b$.
Every feasible completion under the node constraints must satisfy
\[
  \bigvee_{b\in\mathcal B}(x_b=0)
  \quad\Longleftrightarrow\quad
  \sum_{b\in\mathcal B}x_b\le |\mathcal B|-1.
\]
This is a nogood constraint: the blocking reservations cannot all be retained together, as in logic-based and combinatorial Benders feedback~\cite{hooker2003logic,codato2006combinatorial}.
FOCBS enforces it through CBS children, each imposing one $x_b=0$; these children may overlap.
A minimum-capacity cut need not yield an inclusion-minimal or minimum-cardinality blocker set.

Fig.~\ref{fig:flow_conflict} illustrates this process.
At the root, $a_1$ enters $v_3$ and $a_2$ enters $v_4$ at $t=1$.
The unassigned agent $u_1$ at $v_3$ can neither stay nor move to $v_1$ or $v_4$, and moving to $v_6$ only postpones the conflict because $a_1$ and $a_2$ reach $v_6$ and $v_7$ at $t=2$.
Thus, no U-paths exist although $\sst=4$. The min-cut includes the two vertex blockers shown in the figure, $(a_1,v_3,1)$ and $(a_2,v_4,1)$.
In child $N_{b1}$, $a_1$ waits one step at $v_1$, $u_1$ moves to $v_4$ after $a_2$ leaves it, and the oracle returns a plan with $\sst=5$ and $\smu=1$.

With $m=0$, every oracle call succeeds trivially, and FOCBS reduces to CBS over assigned agents.
CBM~\cite{ma2016cbm} also combines CBS with min-cost flow, but its teams are routed to prescribed target sets; FOCBS lets targetless agents stop at any unreserved vertex and turns their routing failures into constraints on assigned agents.

\subsection{BZ-SSP Acceleration}
\label{sec:bzssp}

Under fixed A-paths, an unassigned agent whose start is never reserved can wait in place at zero cost, yet plain SSP still spends one shortest-path search on it.
Batched zero-cost SSP (BZ-SSP) exploits these unreserved starts, as illustrated in Fig.~\ref{fig:flow_oracle_ten}(b).
Algorithm~\ref{alg:bzssp} first inserts zero-cost wait-in-place flows for these starts, then runs SSP only for the remaining flow units on the residual TEN.

\begin{algorithm}[t]
\caption{BZ-SSP Routing Subroutine.}
\label{alg:bzssp}
\begin{algorithmic}[1]
\STATE \textbf{Input:} TEN $\mathcal{N}(\pi^A,T_N)$, starts $\hat S=\{\hat{s}_j\}_{j=1}^m$
\STATE \textbf{Output:} U-path set $\pi^U$, or \textsc{FlowInfeasible}
\STATE Add zero-cost waits for each $\hat{s}_j$ unreserved on $[0,T_N]$
\STATE $k_0 \leftarrow$ number of inserted paths; potentials $\phi_v \leftarrow 0$
\FOR{$i=1$ \TO $m-k_0$}
  \STATE $P \leftarrow$ shortest $s^+$--$s^-$ path under reduced costs $c^\phi$
  \STATE \textbf{if} $P$ does not exist \textbf{then return} \textsc{FlowInfeasible}
  \STATE Update potentials; augment along $P$
\ENDFOR
\RETURN Decompose the value-$m$ flow as $\pi^U$
\end{algorithmic}
\end{algorithm}

In Algorithm~\ref{alg:bzssp}, \textsc{FlowInfeasible} means that the given TEN has no flow of value $m$; $\textsc{FlowOracle}$ then extracts blockers from the min-cut as described in Section~\ref{sec:flow_conflict}.
Phase~1 (line~3) inserts a zero-cost partial flow, after which Phase~2 (lines~5--9) runs SSP for the remaining $m{-}k_0$ units on the full residual TEN, allowing warm-started paths to be rerouted; Proposition~\ref{prop:bzssp} establishes optimality.
During TEN construction, we mark each vertex reserved at any time on $[0,T_N]$. With this reservation lookup, Phase~1 takes $O(m+k_0T_N)$ time and replaces $k_0$ Dijkstra searches. In Phase~2, each Dijkstra search terminates early once $s^-$ is settled.

\subsection{Theoretical Analysis}
\label{sec:theory}

\begin{proposition}[CT cost and horizon monotonicity]
\label{prop:ct_monotonicity}
Suppose each CT node stores, for every assigned agent, a path with minimum last-arrival time under that agent's constraints.
Each child inherits its parent's constraints and adds a new constraint on an assigned agent.
Along any CT branch, the assigned agents' last-arrival times, $\sst$, and $T_N$ are nondecreasing.
Every node also satisfies $T_N\le\sst(\pi_N^A)$.
\end{proposition}
\begin{proof}
Each A-path stored at $N$ has minimum last-arrival time under its constraints. Thus, for any A-path set $\bar\pi^A$ satisfying the constraints at $N$,
\begin{equation}
\begin{aligned}
t_i^{\mathrm{last}}(\bar\pi_i^A)
&\ge t_i^{\mathrm{last}}(\pi_{N,i}^A),
&& \forall a_i\in A,\\
T(\bar\pi^A)&\ge T_N.
\end{aligned}
\label{eq:horizon_invariant}
\end{equation}
The second inequality follows by taking the maximum over agents in the first.
A child's paths also satisfy its parent's constraints; therefore,~\eqref{eq:horizon_invariant} shows that each assigned agent's last-arrival time and the horizon cannot decrease from parent to child.
Summing the last-arrival times shows that $\sst$ cannot decrease either.
Finally, $T_N$ is the maximum of these nonnegative times and $\sst(\pi_N^A)$ is their sum; hence $T_N\le\sst(\pi_N^A)$.
\end{proof}

All A-paths wait after their last arrival, and unassigned agents may stop at any unreserved vertex.
Consequently, for every feasible complete plan $\bar\pi$ whose A-paths satisfy the constraints at $N$, the restriction of $\bar\pi$ to time steps $0,\ldots,T_N$ is conflict-free and induces a value-$m$ flow on the skeleton over layers $0,\ldots,T_N$, with capacities determined by $\bar\pi^A$ on these time steps.
Reservations and waits after $T_N$ lie outside these layers and cannot change the capacity of any cut in them.

\begin{lemma}[Oracle correctness]
\label{lem:oracle}
For fixed conflict-free A-paths $\pi^A\in\mathcal A$, the flow oracle returns U-paths in $\mathcal R^*(\pi^A)$, or correctly reports $\mathcal R(\pi^A)=\emptyset$.
\end{lemma}
\begin{proof}
The construction in Section~\ref{sec:ten} reduces U-path planning to a min-cost flow problem whose minimum cost equals the minimum $\smu$.
Integral capacities guarantee an integral optimum~\cite{ahuja1993network}, whose canonical decomposition lies in $\mathcal R^*(\pi^A)$. If the flow problem has no solution, then $\mathcal R(\pi^A)=\emptyset$.
\end{proof}

\begin{figure}[!t]
  \centering
  \begingroup
  \setlength{\fboxsep}{0pt}
  \setlength{\fboxrule}{0.2pt}
  \begin{minipage}[b]{0.14\columnwidth}\centering
    \fbox{\includegraphics[width=0.94\linewidth]{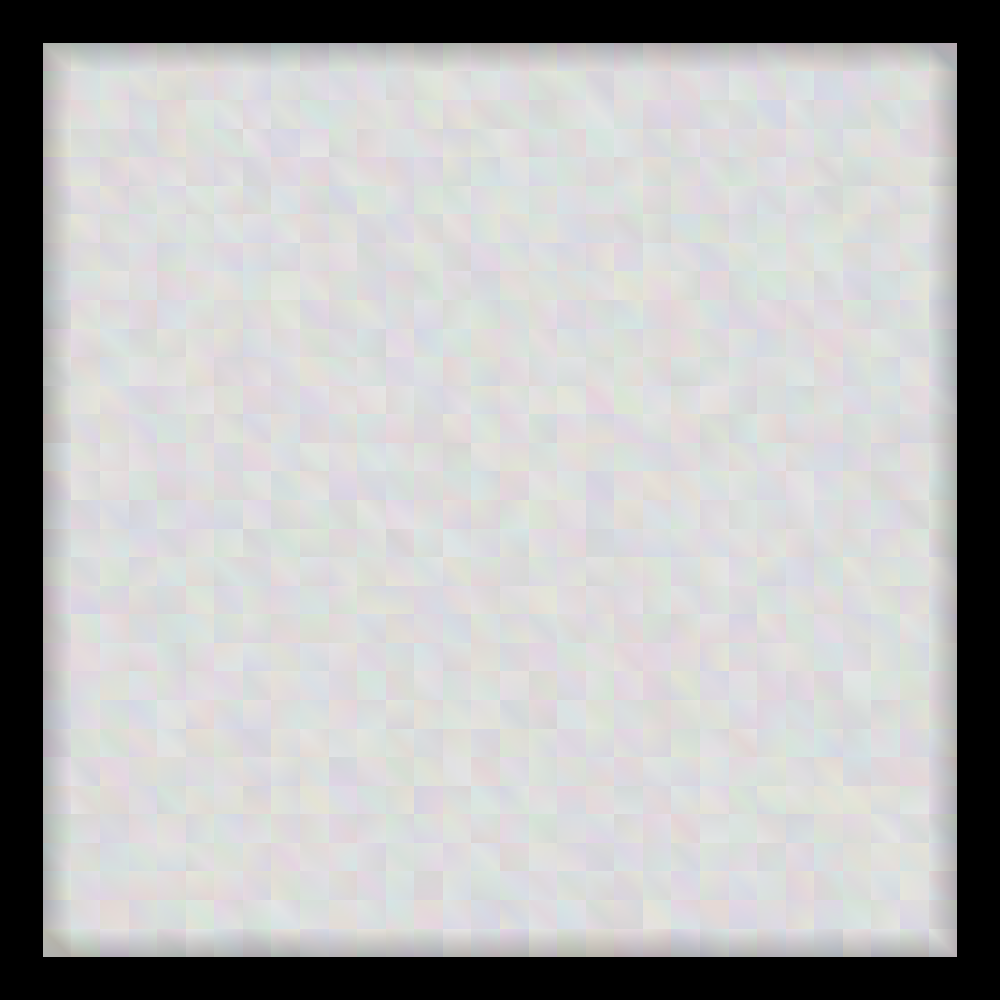}}\\[1pt]
    \subfigtext{(a)}
  \end{minipage}\hfill
  \begin{minipage}[b]{0.14\columnwidth}\centering
    \fbox{\includegraphics[width=0.94\linewidth]{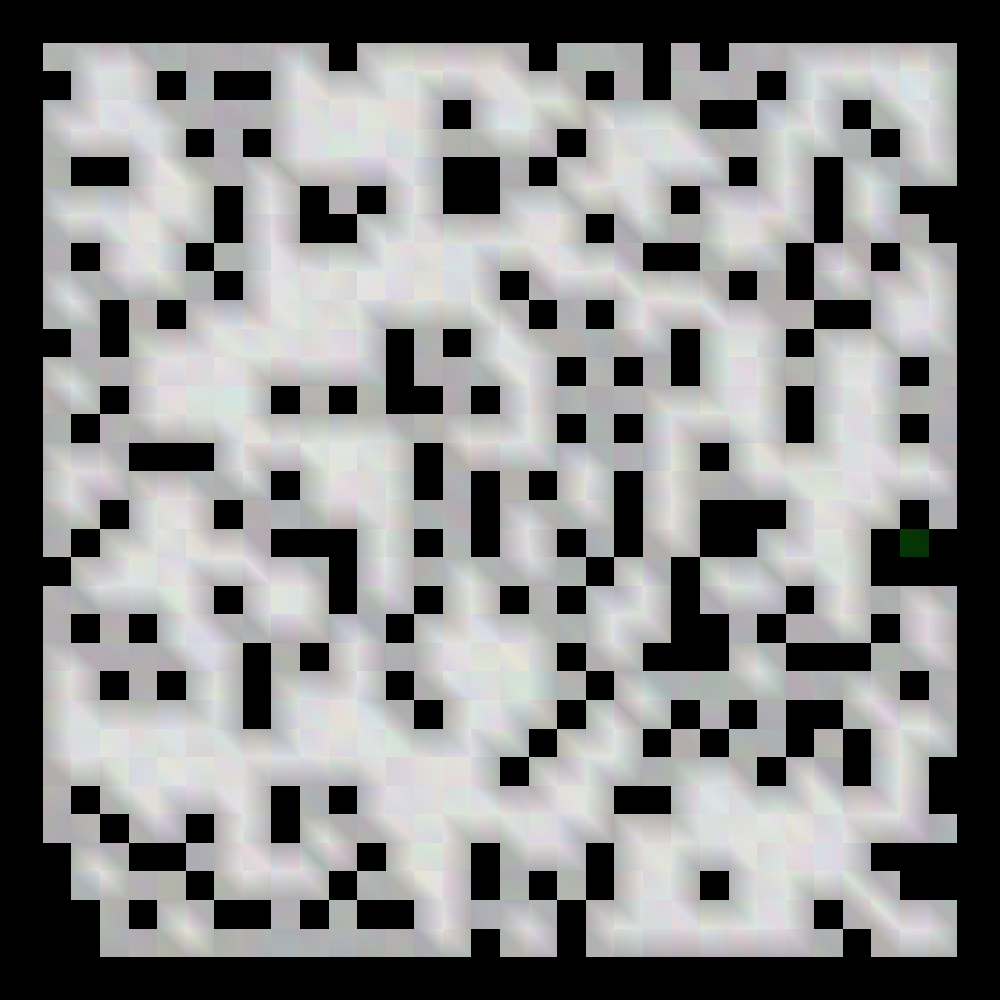}}\\[1pt]
    \subfigtext{(b)}
  \end{minipage}\hfill
  \begin{minipage}[b]{0.14\columnwidth}\centering
    \fbox{\includegraphics[width=0.94\linewidth]{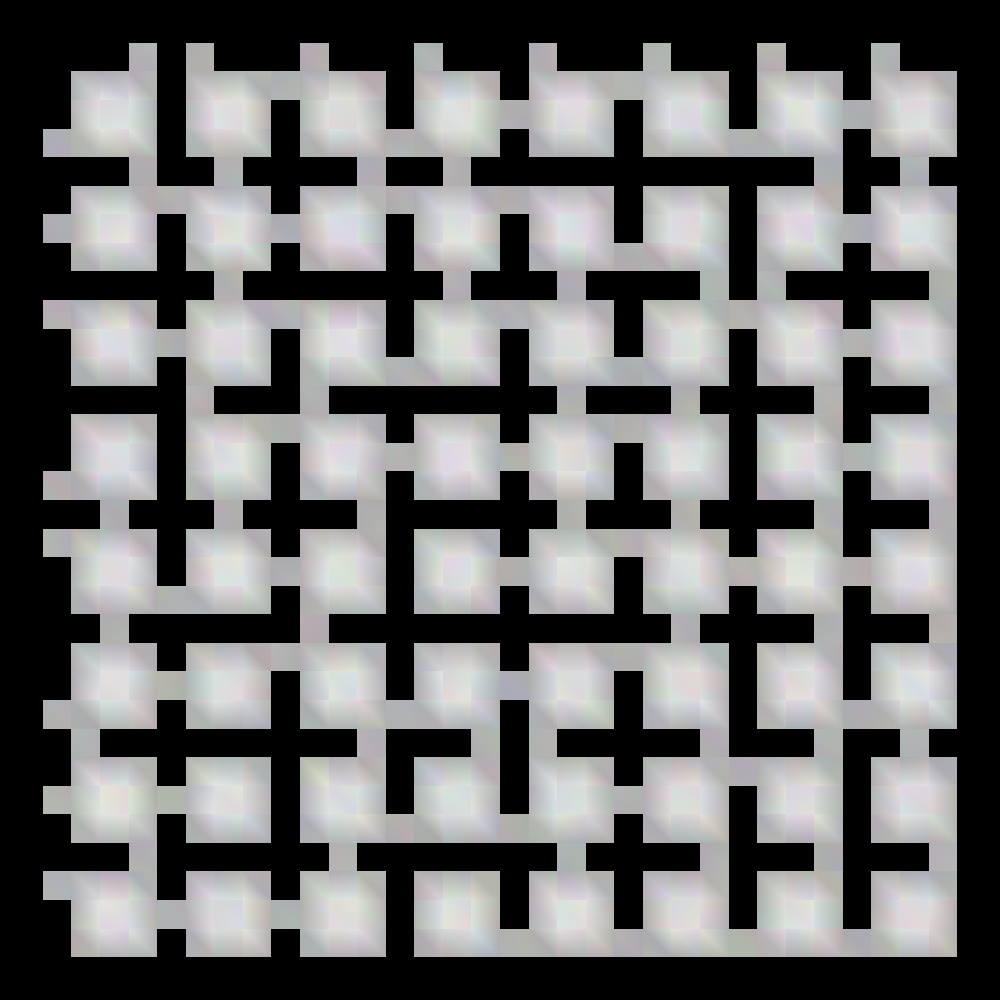}}\\[1pt]
    \subfigtext{(c)}
  \end{minipage}\hfill
  \begin{minipage}[b]{0.14\columnwidth}\centering
    \fbox{\includegraphics[width=0.94\linewidth]{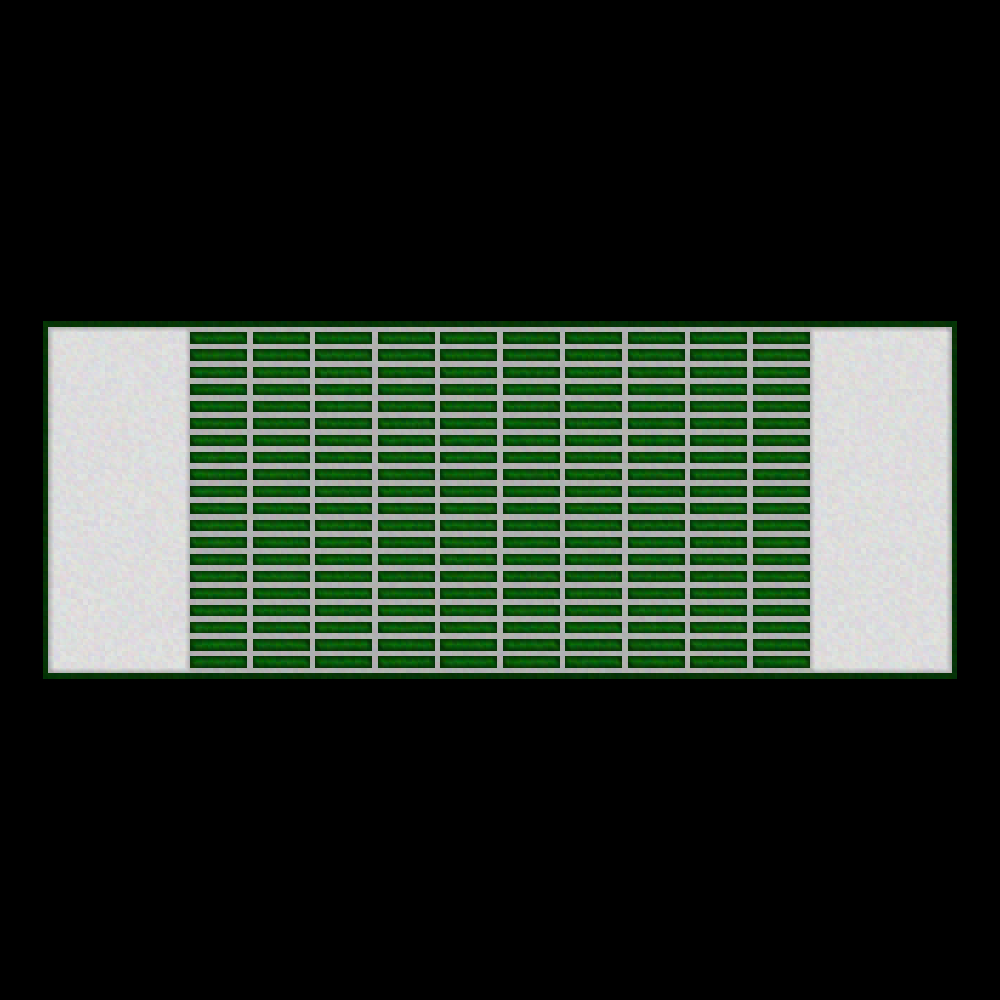}}\\[1pt]
    \subfigtext{(d)}
  \end{minipage}\hfill
  \begin{minipage}[b]{0.14\columnwidth}\centering
    \fbox{\includegraphics[width=0.94\linewidth]{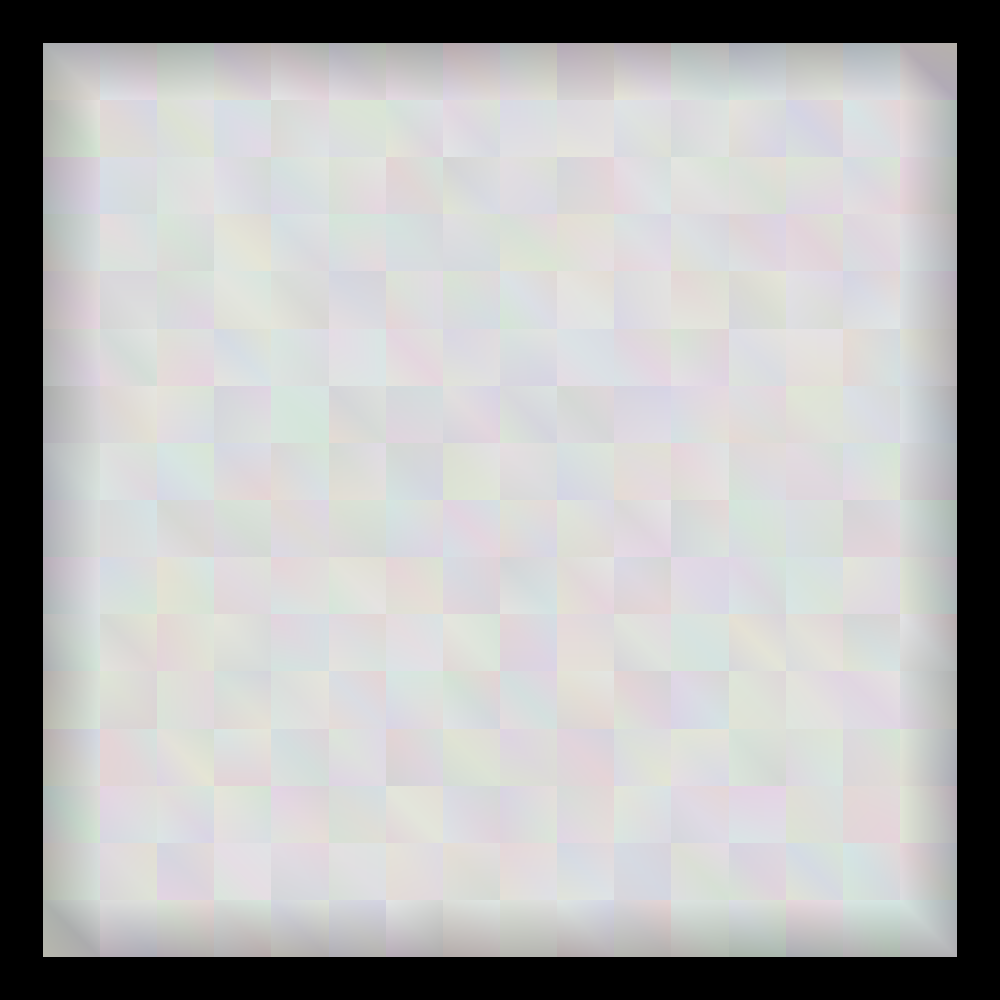}}\\[1pt]
    \subfigtext{(e)}
  \end{minipage}\hfill
  \begin{minipage}[b]{0.24\columnwidth}\centering
    \fbox{\includegraphics[width=0.94\linewidth]{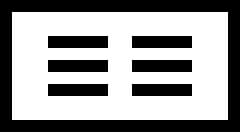}}\\[1pt]
    \subfigtext{(f)}
  \end{minipage}
  \endgroup
  \caption{Six experimental maps: (a) empty-32-32, an open
  32$\times$32 grid; (b) random-32-32-20, a 32$\times$32 grid with 20\%
  random obstacles; (c) room-32-32-4, a 32$\times$32 grid with room
  partitions at four-cell intervals; (d) warehouse-10-20-10-2-1, containing a
  10$\times$20 array of 10$\times$2 shelf blocks with one-cell internal
  aisles; (e) empty-16-16, an open 16$\times$16 grid; and
  (f) warehouse-small-20-11, a custom 20$\times$11 warehouse with six
  five-cell shelf blocks.}
  \label{fig:maps}
\end{figure}

\begin{proposition}[BZ-SSP optimality preservation]
\label{prop:bzssp}
If a flow of value $m$ exists in the given TEN, BZ-SSP returns one of minimum cost; otherwise, it reports that no such flow exists.
\end{proposition}
\begin{proof}
The initial $k_0$ wait paths form a feasible zero-cost flow, which is minimum-cost for value $k_0$ because all original arc costs are nonnegative.
Their reverse arcs also have zero cost; thus all initial residual arc costs are nonnegative.
SSP continues on the full residual network, allowing rerouting and preserving minimum cost at each flow value~\cite{ahuja1993network}.
If SSP finds no augmenting path before reaching $m$, the required flow value cannot be achieved.
\end{proof}

\begin{figure*}[!t]
  \centering
  \includegraphics[width=0.90\textwidth]{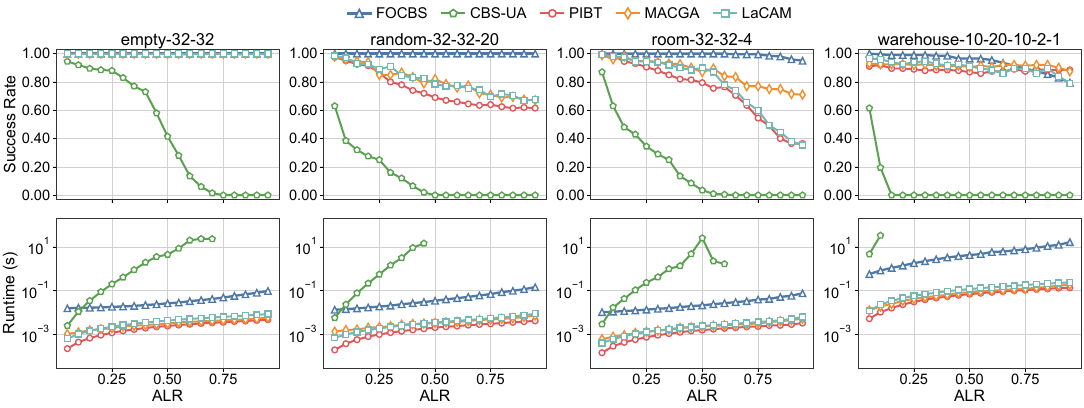}
  \caption{Success rate and median runtime over solved instances as
  unassigned-agent load increases; runtime is shown on a logarithmic scale.}
  \label{fig:scalability_revised}
\end{figure*}

\begin{lemma}[Blocker validity]
\label{lem:blocker}
Consider a CT node $N$ with mutually conflict-free A-paths $\pi_N^A\in\mathcal A$. Suppose no conflict-free U-paths exist under these A-paths, i.e., $\mathcal R(\pi_N^A)=\emptyset$. For every assigned agent, the path stored at $N$ satisfies the node constraints and has minimum last-arrival time under those constraints. Let $S$ be the source-reachable side after max flow. Let $\mathcal B$ be the branchable labeled blockers in $\delta^+(S)$. Every feasible complete plan whose A-paths satisfy the current branch constraints releases at least one blocker in $\mathcal B$. Thus branching on all $b\in\mathcal B$ is a valid disjunction. If $\mathcal B=\emptyset$, no feasible complete plan has A-paths satisfying the constraints at $N$.
\end{lemma}
\begin{proof}
Let $c_N$ be the capacities at $N$. Since the TEN under $\pi_N^A$ has no flow of value $m$, $c_N(\delta^+(S))<m$.
For contradiction, suppose a feasible complete plan $\bar\pi$ whose A-paths satisfy the constraints at $N$ retains every blocker in $\mathcal B$.
By the horizon invariant~\eqref{eq:horizon_invariant}, $T(\bar\pi^A)\ge T_N$.
The restriction property stated above therefore implies that $\bar\pi^U|_{0:T_N}$ induces a value-$m$ flow on the skeleton over layers $0,\ldots,T_N$, with capacities $\bar c$ determined by $\bar\pi^A|_{0:T_N}$.
Hence $\bar c(\delta^+(S))\ge m$.

Now compare the capacities of the arcs in $\delta^+(S)$.
Structural capacities are independent of the A-paths, and an arc already having unit capacity under $c_N$ cannot gain capacity.
The $t{=}0$ vertex reservations are fixed by the initial configuration.
Therefore, a cut arc can gain capacity only if a labeled zero-capacity reservation at $N$ is released.
Every such releasable reservation belongs to $\mathcal B$: fixed $t{=}0$ vertex reservations have been discarded, and a reservation already forbidden by a branch constraint cannot occur in $\pi_N^A$.
Since $\bar\pi$ retains every blocker in $\mathcal B$, no cut arc gains capacity, and thus $\bar c(\delta^+(S))\le c_N(\delta^+(S))$. Hence
\[
 m\le \bar c(\delta^+(S))\le c_N(\delta^+(S))<m,
\]
a contradiction. Thus some blocker must be released, proving the disjunction and the case $\mathcal B=\emptyset$.
\end{proof}

The following termination result applies only to feasible instances. The value $C^*$ serves only as an existential bound in the proof and need not be known or used by FOCBS; when $\mathcal A_{\mathrm{comp}}=\emptyset$, FOCBS is not guaranteed to terminate.

\begin{theorem}[Sequential optimality of FOCBS]
\label{thm:optimality}
For a MAPF-UA instance on a finite graph, if $\mathcal A_{\mathrm{comp}}\neq\emptyset$, FOCBS terminates and returns a sequentially optimal solution in the sense of Definition~\ref{def:seq_mapfua}.
For the returned solution $\hat\pi$, the $\smu$ guarantee is conditional on $\hat\pi^A$: FOCBS guarantees $\smu(\hat\pi)=Q(\hat\pi^A)$, but not $\hat\pi^A\in\arg\min_{\pi^A\in\mathcal A^*}Q(\pi^A)$.
Thus, it does not guarantee global lexicographic optimality.
\end{theorem}
\begin{proof}
Fix $\pi^{*A}\in\mathcal A^*$ and any $\pi^{*U}\in\mathcal R(\pi^{*A})$, and let $\pi^*=(\pi^{*A},\pi^{*U})$.
Call a CT node compatible when $\pi^{*A}$ satisfies its constraints; the root is compatible.
At every compatible node, applying the horizon invariant~\eqref{eq:horizon_invariant} with $\bar\pi^A=\pi^{*A}$ gives $t_i^{\mathrm{last}}(\pi_{N,i}^A)\le t_i^{\mathrm{last}}(\pi_i^{*A})$ for every $a_i$, and hence $\sst(\pi_N^A)\le C^*$ and $T_N\le T(\pi^{*A})\le C^*$.
A standard CBS split preserves at least one compatible child.
At a node with $\pi_N^A\in\mathcal A$ and $\mathcal R(\pi_N^A)=\emptyset$, Lemma~\ref{lem:blocker}, applied to $\pi^*$, shows at least one flow-conflict child remains compatible.
Until a solution is returned, OPEN contains a compatible node of cost at most $C^*$.

The compatible-node invariant ensures that best-first search never expands a node with key greater than $C^*$ before returning a solution.
Although horizons may grow along a CT branch, every node expanded before that return satisfies $T_N\le\sst(\pi_N^A)\le C^*$.
Therefore, every generated vertex or edge constraint refers to an agent, graph element, and time step in a finite set bounded by $C^*$.
Since every child adds a new constraint, only finitely many nodes of cost at most $C^*$ can be generated.
This finite set cannot be expanded indefinitely; thus, the search must return after finitely many expansions a node $N$ with $\pi_N^A\in\mathcal A$, $\mathcal R(\pi_N^A)\neq\emptyset$, and $\sst(\pi_N^A)\le C^*$.
Feasibility implies $\pi_N^A\in\mathcal A_{\mathrm{comp}}$. With $\sst(\pi_N^A)\le C^*$ and the definition of $C^*$, this gives $\sst(\pi_N^A)=C^*$ and hence $\pi_N^A\in\mathcal A^*$. Lemma~\ref{lem:oracle} gives $\pi_N^U\in\mathcal R^*(\pi_N^A)$, establishing sequential optimality.
\end{proof}

\section{Experiments}
\label{sec:experiments}

The scalability experiment increases unassigned-agent load, and the
fixed-density experiment varies the proportions of assigned agents and
unassigned agents. Ablation studies evaluate flow-conflict branching,
BZ-SSP, and tie-breaking, followed by simulation and physical-robot validation.

\begin{figure*}[!t]
  \centering
  \includegraphics[width=0.90\textwidth]{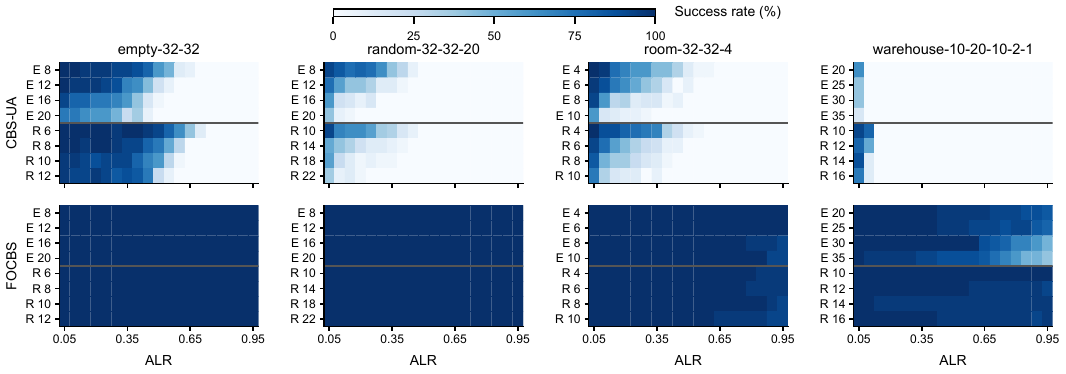}\\[-0.2ex]
  \subfigtext{(a)}\\[0.6ex]
  \includegraphics[width=0.90\textwidth]{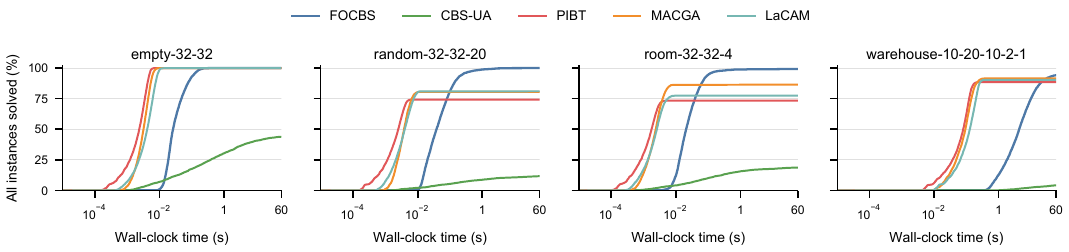}\\[-0.2ex]
  \subfigtext{(b)}
  \caption{Scalability results by assigned-agent load and time budget. (a) Success rate by
  scenario type (E: \emph{even}; R: \emph{random}) and assigned-agent count.
  (b) Cumulative percentage of all 3,800 instances per map solved within
  each wall-clock time budget; failed instances remain in the denominator,
  and the time axis is logarithmic.}
  \label{fig:broad_stratified}
\end{figure*}

\subsection{Experimental Settings}
\label{sec:exp_settings_revised}

\subsubsection{Instances}
Fig.~\ref{fig:maps} shows the four standard MAPF maps used in the scalability experiment~\cite{stern2019mapf} (an open grid, random obstacles,
bottleneck rooms, and warehouse aisles) and the smaller open grid and custom
warehouse used in the fixed-density experiment.  Assigned-agent start--target
pairs come from the standard \emph{even} and \emph{random} scenarios, and
unassigned-agent starts are sampled without replacement from the remaining
traversable vertices.  All solvers receive the same seeded instances.

\subsubsection{Methods}
The scalability experiment compares FOCBS with CBS-UA and the fast yielding baselines
PIBT, MACGA, and LaCAM~\cite{sharon2015cbs,okumura2022pibt,pertzovsky2025cga,okumura2024engineeringlacam}.
CBS-UA runs CBS on all agents, and each unassigned agent may stop at any
vertex that satisfies its constraints. CBS-UA minimizes $\sst$ but does not optimize $\smu$.
For PIBT, MACGA, and LaCAM, unassigned agents wait by default and yield with the
lowest priority. The yielding baselines do not guarantee sequential optimality.
The fixed-density experiment additionally distinguishes
CBS-static, which fixes unassigned agents as obstacles; LCBS, our joint
lexicographic CBS baseline for MAPF-UA, which adapts the lexicographic bi-objective
framework of~\cite{zhao2025lcbs}, places all agents in one constraint tree, and globally
minimizes $(\sst,\smu)$; CBS-SSP, a two-stage
baseline with the same sequential objective as FOCBS; and FOCBS-noTB,
which removes only the start-avoiding tie breaker.
CBS-SSP first obtains $\sst$-optimal A-paths with CBS$_0$, which runs CBS on assigned agents only, and then calls an SSP oracle once to minimize $\smu$.
CBS-SSP stops if no U-paths exist.

\subsubsection{Metrics and Platform}
We report success rate, median runtime over solved instances (interpreted
jointly with success rate), $\sst$, and $\smu$.  Objective values are compared on
map-specific common solved sets (Table~\ref{tab:fixed_alr_objectives}). CBS-static is excluded from the
objective-value comparison because fixing unassigned agents forces $\smu=0$. Benchmarks
use Ubuntu~24.04 with an Intel Core i9-14900K and 64~GB RAM.
All methods are implemented in C++ and run single-threaded; PIBT, MACGA, and LaCAM are our reimplementations adapted to MAPF-UA.  The scalability
and fixed-density experiments use a 60-s limit; the BZ-SSP ablation below uses a
120-s limit to retain enough jointly solved warehouse pairs.

\subsection{Scalability with Increasing Unassigned-Agent Load}
\label{sec:exp_scalability_revised}

We fix the assigned-agent count $n$ and add unassigned agents to increase
$\alr$ from 0.05 to 0.95 on the four standard maps. Optimal CBS becomes
rapidly more expensive as $n$ increases. To keep assigned-agent search
tractable and isolate the effect of the unassigned-agent count $m$, we
select four values of $n$ per map and scenario type for which CBS$_0$
solves all 25 seeds with $m=0$ within a 10-s per-instance limit. Each map--$\alr$
setting combines the four assigned-agent counts with 25 \emph{even} and
25 \emph{random} seeds, giving 200 instances per method and 76,000 runs
in total.

CBS-UA places both agent groups in one constraint tree, so adding
unassigned agents increases the burden of conflict resolution. Aggregated
over all tested loads, the success rate of CBS-UA is at most 43.9\% on any map
(Fig.~\ref{fig:scalability_revised}). FOCBS confines CBS search to assigned
agents and routes unassigned agents through the flow oracle, achieving
overall success rates of 94.1--100\% across the four maps. The median
runtime of FOCBS over solved instances also grows more gradually with unassigned-agent
count.

PIBT, MACGA, and LaCAM return solutions faster than FOCBS, but achieve
lower overall success rates on the three obstacle maps (73.4--91.5\%
across the tested loads; Figs.~\ref{fig:scalability_revised}
and~\ref{fig:broad_stratified}(b)). On instances solved by both methods, their $\sst$ exceeds that of FOCBS by 2.0--27.0\%, and their mean $\smu$ is 2.9--5.9 times that of FOCBS. Under the tested assigned-agent loads,
FOCBS finds globally $\sst$-optimal solutions within 1~s for the vast
majority of instances on the empty, random, and room maps, whereas the
warehouse map requires longer runtimes.

Fig.~\ref{fig:broad_stratified}(a) shows that the success rate of CBS-UA
declines with increasing unassigned-agent count at every tested
assigned-agent count. FOCBS maintains higher success rates under the
corresponding settings, showing that the advantage over joint search
holds across different assigned-agent loads. In the random warehouse
scenarios, FOCBS solves all 25 instances with 10 assigned agents and
5,404 unassigned agents, indicating that FOCBS scales well with the
number of unassigned agents.

On the warehouse map, the success rate of FOCBS begins to decline at
lower densities as the assigned-agent count increases. Assigned-agent
paths reserve space-time resources in the TEN, restricting the space
available for unassigned agents to yield. Even when the assigned-agent
load keeps CBS search tractable, increasing the number of assigned agents
can make unassigned-agent routing through the TEN more difficult.

\subsection{Effect of Agent Composition at Fixed Density}
\label{sec:exp_composition_revised}

The composition experiment fixes $\alr=0.25$ on empty-16-16 and warehouse-small-20-11
($n+m=64$ and 33).  With starts fixed, unassigned agents
are reassigned in steps of five. The experiment covers 20 compositions and six
methods, totaling 6,000 runs.  Because CBS-based search becomes rapidly harder as the number of agents grows,
small maps and relatively low density allow all methods to
solve enough common instances for quality comparison.

\begin{figure}[!t]
  \centering
  \includegraphics[width=0.90\columnwidth]{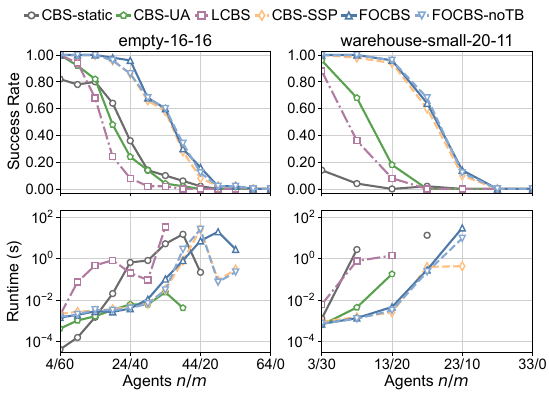}
  \caption{Success rate (top) and median runtime over solved instances
  (bottom, log scale) at ALR = 0.25. Each setting has 25 \emph{even} and
  25 \emph{random} instances; $n/m$ denotes the numbers of assigned agents and unassigned agents, respectively.
  No runtime is plotted when none are solved.}
  \label{fig:fixed_composition}
\end{figure}

Fig.~\ref{fig:fixed_composition} shows that FOCBS substantially outperforms
CBS-UA and LCBS when unassigned agents predominate. At warehouse
$n/m=13/20$, FOCBS solves 48 of 50 instances, while CBS-UA and LCBS solve
at most nine. FOCBS exploits the asymmetry between the two agent types
by routing unassigned agents through the flow oracle, reducing the
combinatorial burden on CBS. As the proportion of assigned agents
increases, CBS search becomes a common bottleneck for all three methods,
making instances harder to solve and narrowing the performance gap.

The low success rate of CBS-static in the small warehouse highlights
the need to let unassigned agents yield in narrow aisles. CBS-SSP also
benefits from separating the two agent groups and nearly matches FOCBS
at $\alr=0.25$, where the first A-path set is usually completable;
Section~\ref{sec:exp_components_revised} evaluates flow-conflict branching
under higher loads.

\begin{table*}[!t]
  \centering
  \caption{Solution Quality on Common Solved Sets}
  \label{tab:fixed_alr_objectives}
  \setlength{\tabcolsep}{4.2pt}
  \renewcommand{\arraystretch}{1.05}
  \footnotesize
  \begin{tabular*}{\textwidth}{@{\extracolsep{\fill}}lccrrrrr@{}}
    \toprule
    Map & Common set & Common mean $\mathrm{SST}$ &
    \multicolumn{5}{c}{Mean $\mathrm{SMU}$} \\
    \cmidrule(lr){4-8}
        & & & CBS-UA & LCBS & CBS-SSP & FOCBS & FOCBS-noTB \\
    \midrule
    empty-16-16 & 145/650 & 106.30 & 5.19 & \textbf{4.39} & 20.21 & 6.14 & 20.21 \\
    warehouse-small-20-11 & 64/350 & 54.86 & 8.50 & \textbf{7.06} & 13.88 & 9.00 & 13.88 \\
    \midrule
    empty-32-32 & 1668/3800 & 226.82 & 20.35 & \textemdash & \textemdash & \textbf{19.56} & 60.66 \\
    warehouse-10-20-10-2-1 & 162/3800 & 1236.12 & 50.93 & \textemdash & \textemdash & \textbf{49.01} & 171.58 \\
    \bottomrule
  \end{tabular*}
  \vspace{2pt}\par
  \parbox{\textwidth}{\footnotesize Rows 1--2 use the five-method
  fixed-density common sets; rows 3--4 use the CBS-UA/FOCBS/FOCBS-noTB common
  sets from the scalability experiment. Bold marks the lowest
  mean $\mathrm{SMU}$. Common sets are biased toward easier instances. CBS-SSP and FOCBS-noTB return the same A-paths and $\mathrm{SMU}$.}
\end{table*}

The comparison with LCBS evaluates the tradeoff between the sequential
and global lexicographic objectives. Both methods attain the same
optimal $\sst$ on the common solved sets in
Table~\ref{tab:fixed_alr_objectives}. FOCBS solves 2.4 times as many
instances, and the median paired runtime ratio
$t_{\mathrm{LCBS}}/t_{\mathrm{FOCBS}}$ is 25.7 on instances solved by both
methods. LCBS achieves lower mean $\smu$ because it searches over
alternative $\sst$-optimal assigned-agent paths; FOCBS minimizes $\smu$
for the selected assigned-agent paths. The results support the practical
value of the sequential objective: FOCBS retains globally optimal
$\sst$ while accepting additional unassigned-agent
movement for faster solving and higher success rates.

CBS-UA also attains lower mean $\smu$ on the two small maps because its joint search tends to select A-paths that avoid unassigned agents; on all four maps of the scalability experiment, its mean $\smu$ exceeds that of FOCBS.

\subsection{Ablation Experiments and Runtime Analysis}
\label{sec:exp_components_revised}

\subsubsection{Flow-Conflict Branching}

\begin{figure}[t]
  \centering
  \includegraphics[width=0.90\columnwidth]{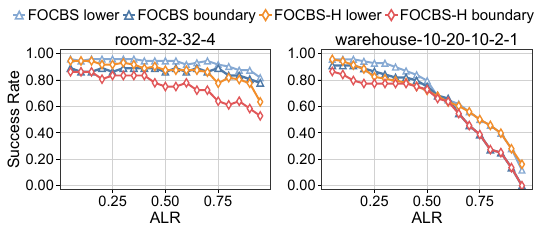}
  \caption{Success rates of FOCBS and FOCBS-H. ``Lower''/``boundary'' denote
  the two smaller/larger assigned-agent loads: $n=\{18,20\}/\{22,24\}$ for room
  (both scenario types), $\{50,55\}/\{60,65\}$ for warehouse-even, and
  $\{45,50\}/\{55,60\}$ for warehouse-random. Only seeds solved by the
  10-s CBS$_0$ probe with only assigned agents ($m=0$, 25 seeds per setting) are plotted.}
  \label{fig:flow_conflict_ablation_revised}
\end{figure}

To isolate flow-conflict branching, we compare FOCBS with FOCBS-H.
When no collision-free U-paths exist under the current A-paths, FOCBS branches and FOCBS-H prunes the node.
Both variants disable the tie breaker and use the same seeds solved by CBS$_0$.

Fig.~\ref{fig:flow_conflict_ablation_revised} shows that flow-conflict
branching improves success on both maps. Scarce free space makes
unassigned-agent routing more sensitive to the selected assigned-agent
paths, so the first conflict-free A-path set often admits no U-paths. Instead of pruning such a node, flow-conflict branching uses the min cut to locate the A-path reservations that block unassigned agents and constrains only those agents, so CBS revises the responsible A-paths without losing $\sst$ optimality. Among the recorded instances whose first oracle call is infeasible, FOCBS recovers a solution within 60~s in 203 of 242 room instances (83.9\%) and 80 of 482 warehouse instances (16.6\%). On room, the gain grows with density; on warehouse, it narrows markedly above $\alr=0.55$, where the success rates of both variants decline together.

On one room-32-32-4 instance ($n=18$, $m=596$), delaying a single A-path by one step resolves the flow conflict; FOCBS solves it in 0.39~s, whereas FOCBS-H fails within 60~s.

\subsubsection{BZ-SSP}

We compare full FOCBS with FOCBS-noBZ, which disables only the BZ-SSP warm
start.  The experiment uses the four standard
maps, $\alr\in\{0.10,0.95\}$, four assigned-agent loads per map, and 25
\emph{even} scenarios per map--ALR--load setting: 800 paired instances with
a 120-s limit for both variants.  Table~\ref{tab:bz_end_to_end} reports median
end-to-end wall-clock runtimes over jointly solved pairs.

\begin{table}[!t]
  \centering
  \caption{Median End-to-End FOCBS Runtimes With and Without BZ-SSP}
  \label{tab:bz_end_to_end}
  \setlength{\tabcolsep}{1.5pt}
  \renewcommand{\arraystretch}{1.05}
  \footnotesize
  \begin{tabular}{@{}p{0.35\columnwidth}ccrrc@{}}
    \toprule
    Map & $\alr$ & Solved & \multicolumn{2}{c}{Time (s)} & Speedup [IQR] \\
    \cmidrule(lr){4-5}
      & & & noBZ & FOCBS & \\
    \midrule
    empty-32-32 & 0.10 & 100/100 & 0.030 & 0.019 & 1.51 [1.43--1.58] \\
     & 0.95 & 100/100 & 0.272 & 0.141 & 1.91 [1.69--2.19] \\
    \addlinespace[1pt]
    random-32-32-20 & 0.10 & 100/100 & 0.020 & 0.016 & 1.27 [1.20--1.33] \\
     & 0.95 & 100/100 & 0.242 & 0.160 & 1.51 [1.34--1.76] \\
    \addlinespace[1pt]
    room-32-32-4 & 0.10 & 100/100 & 0.015 & 0.012 & 1.25 [1.21--1.30] \\
     & 0.95 & 96/98 & 0.149 & 0.090 & 1.65 [1.47--2.08] \\
    \addlinespace[1pt]
    warehouse-10-20-10-2-1 & 0.10 & 99/99 & 4.735 & 1.088 & 3.88 [3.50--4.61] \\
     & 0.95 & 64/65 & 80.23 & 48.55 & 1.72 [1.49--1.98] \\
    \bottomrule
  \end{tabular}
  \vspace{2pt}\par
  \parbox{\columnwidth}{\footnotesize\raggedright Times are medians over jointly solved
  pairs; \emph{Solved} is noBZ/FOCBS out of 100. Speedup is paired
  $t_{\mathrm{noBZ}}/t_{\mathrm{FOCBS}}$; brackets report the interquartile range (IQR).\par}
\end{table}

BZ-SSP accelerates FOCBS on all four maps at both tested densities
(Table~\ref{tab:bz_end_to_end}), with an overall median paired speedup of
\mbox{1.54 [IQR: 1.32--2.02]}. Plain SSP performs a shortest-path search
even for an unassigned agent that can wait at its start. BZ-SSP inserts
zero-cost waiting flows in a batch, eliminating those searches while
allowing subsequent rerouting in the residual network. Both variants
return identical $\sst$ and $\smu$ on every jointly solved pair,
confirming that the measured acceleration preserves solution quality.

\subsubsection{Start-Avoiding Tie-Breaking}

FOCBS and FOCBS-noTB differ only in the start-avoiding tie breaker.
TB reduces mean $\smu$ by 35.1--71.4\% across the four common sets in
Table~\ref{tab:fixed_alr_objectives}, with unchanged $\sst$ and similar
success rates and runtimes in Fig.~\ref{fig:fixed_composition}.
An assigned-agent path that visits an unassigned agent's start forces
that unassigned agent to move. Among equal-cost choices, TB favors paths
with fewer such visits, reducing the need for yielding before the flow
oracle plans unassigned-agent paths.

\subsubsection{Computational Cost}
\label{sec:exp_runtime_attribution_revised}

\begin{table*}[!t]
  \centering\footnotesize
  \caption{FOCBS Search Statistics and Runtime Breakdown in the Scalability Experiment}
  \label{tab:search_runtime}
  \setlength{\tabcolsep}{3pt}
  \renewcommand{\arraystretch}{1.12}
  \begin{tabular*}{\textwidth}{@{\extracolsep{\fill}}lrrrrrrrr@{}}
    \toprule
    & \multicolumn{3}{c}{Search statistics}
    & \multicolumn{5}{c}{Share of cumulative wall-clock time (\%)} \\
    \cmidrule(lr){2-4}\cmidrule(lr){5-9}
    Map & \shortstack{Runs\\(timeouts)}
    & \shortstack{CT expansions\\Median / P95}
    & \shortstack{Oracle calls\\Mean per run}
    & \shortstack{CBS\\search}
    & \shortstack{TEN\\construction}
    & \shortstack{TEN flow\\solving}
    & \shortstack{Flow-conflict\\diagnosis}
    & Other \\
    \midrule
    empty-32-32 & 3,800 (0) & 1 / 12 & 1.000 & 4.0 & 35.3 & 57.5 & 0.0 & 3.3 \\
    random-32-32-20 & 3,800 (0) & 6 / 294 & 1.024 & 46.7 & 10.4 & 36.3 & 5.6 & 1.0 \\
    room-32-32-4 & 3,800 (28) & 2 / 63 & 3.999 & 1.9 & 4.0 & 29.9 & 63.4 & 0.8 \\
    warehouse-10-20-10-2-1 & 3,800 (225) & 2 / 36 & 1.050 & 1.1 & 4.7 & 71.0 & 22.7 & 0.5 \\
    \bottomrule
  \end{tabular*}
  \par\vspace{3pt}
  \parbox{\textwidth}{\footnotesize\raggedright
  Parentheses give timeout counts; P95 is the 95th percentile.
  For each map, time shares are cumulative stage time divided by cumulative
  runtime. CBS search includes conflict detection. TEN construction and flow
  solving exclude diagnosis. Flow-conflict diagnosis includes diagnostic TEN
  construction, maximum-flow solving, and blocker extraction; Other includes
  path extraction and remaining work.\par}
\end{table*}

The near-one mean oracle-call counts on three maps in
Table~\ref{tab:search_runtime} indicate that FOCBS usually completes
unassigned-agent routing with a single call. FOCBS invokes the oracle
only after finding mutually conflict-free assigned-agent paths, avoiding
repeated flow computations during most of the CBS search. Consequently,
even on the random map, where assigned-agent search requires more CT
expansions, only 0.64\% of expansions invoke the oracle.

The additional cost in difficult instances is concentrated in flow
computation and diagnosis. On room-32-32-4, the 28 timed-out runs account
for 64.0\% of oracle calls because infeasibility repeatedly triggers
assigned-agent replanning and TEN reconstruction. On the warehouse map,
the timed-out runs contribute only 9.8\% of oracle calls but 93.2\% of
diagnostic time, indicating
expensive diagnoses rather than frequent calls. At $\alr=0.95$ ($m\approx5{,}400$), increasing $n$ from 20 to 35 in \emph{even} scenarios raises the median flow-solving time from 28.0 to 48.6~s, whereas the median CBS time stays below 0.3~s.

\begin{figure}[t]
  \centering
  \includegraphics[width=0.90\columnwidth]{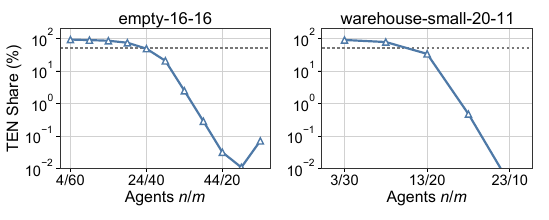}
  \caption{Median TEN share of FOCBS runtime at fixed ALR = 0.25.
  For each solved instance, $t_{\mathrm{TEN}}/(t_{\mathrm{CBS}}+t_{\mathrm{TEN}})$
  includes all oracle construction, solving, and extraction work.
  The dashed line marks 50\%.}
  \label{fig:focbs_ten_share}
\end{figure}

Fig.~\ref{fig:focbs_ten_share} supports the efficiency of routing
unassigned agents through the flow oracle. As the proportion of assigned
agents increases at fixed density, CBS accounts for more runtime while
flow computation remains inexpensive on the tested small maps.
For solved empty-16-16 instances, from $n/m=4/60$ to $34/30$, median CBS
time rises from 0.11 to 103.22~ms, while median TEN time increases by less
than 1~ms. The runtime growth mainly comes from the larger assigned-agent
search. In one random-32-32-20 instance, CBS plans 12 assigned agents in
0.173~s, while a single TEN call routes 766 unassigned agents in 0.174~s. On the other maps of the scalability experiment, flow computation instead dominates runtime (Table~\ref{tab:search_runtime}).

\subsection{Execution Feasibility and Fleet Efficiency}
\label{sec:verification}

We validate FOCBS path executability in a ROS~2/Gazebo warehouse simulation
and on three robots in a chemical laboratory. For each batch, FOCBS plans
paths for assigned robots and unassigned robots. An action dependency graph
(ADG)~\cite{honig2019adg} converts these plans into asynchronous scheduling
constraints, while the ROS navigation stack handles continuous navigation.
Robustness to localization errors and motion uncertainty is beyond this study.

\begin{figure}[!t]
  \centering
  \begin{minipage}{\columnwidth}
    \centering
    \begin{minipage}[t]{0.54\linewidth}\centering
      \includegraphics[width=\linewidth]{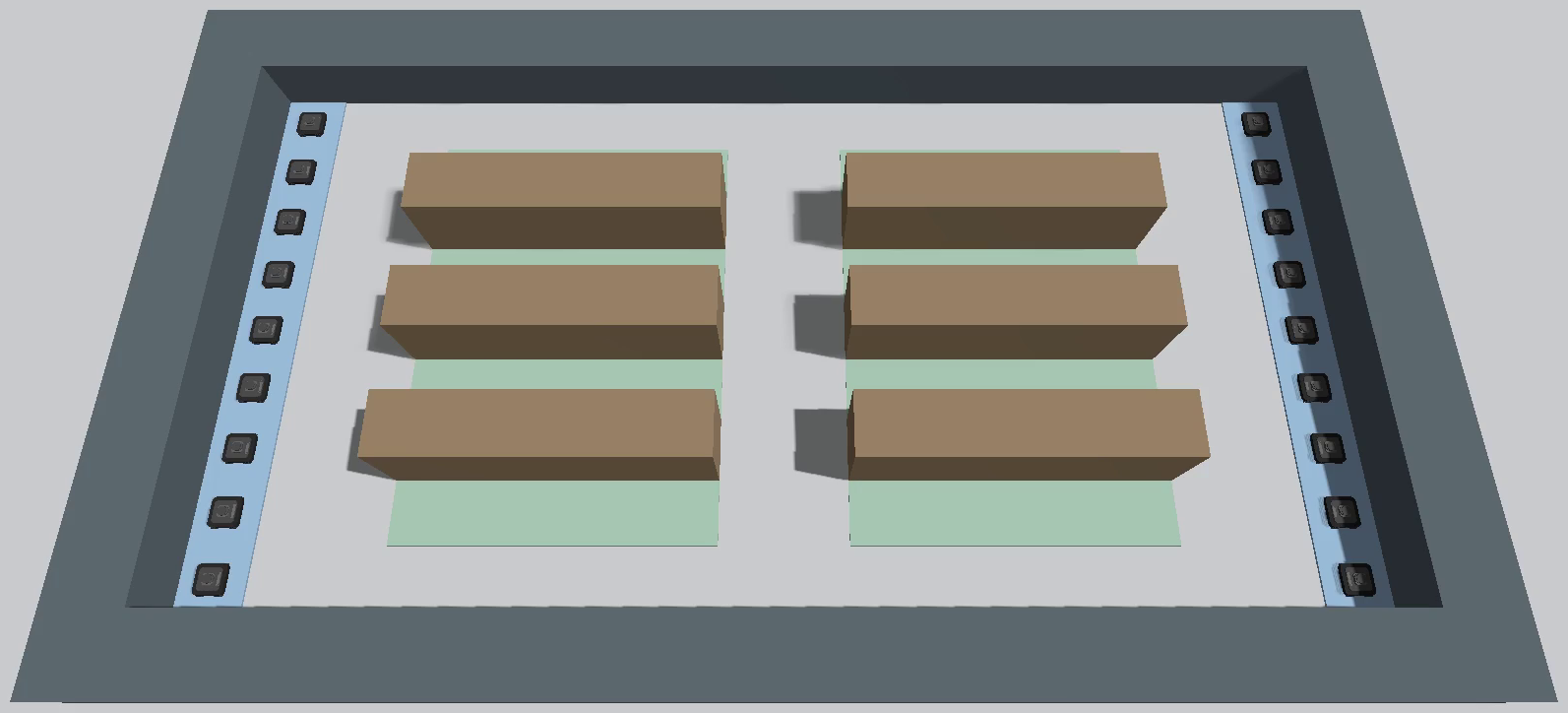}\\[-0.2ex]
      \subfigtext{(a)}
    \end{minipage}\hfill
    \begin{minipage}[t]{0.435\linewidth}\centering
      \includegraphics[width=\linewidth]{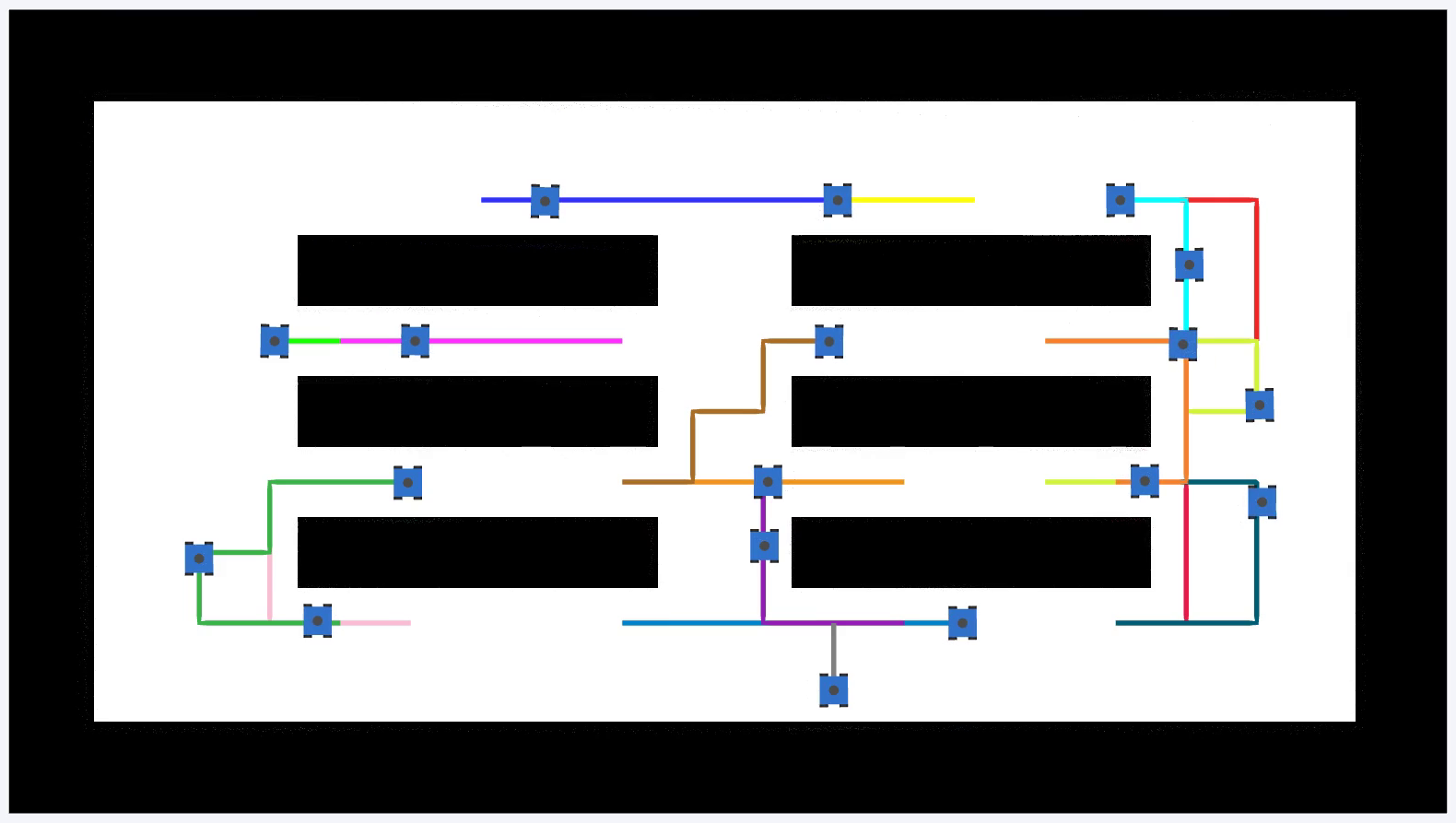}\\[-0.2ex]
      \subfigtext{(b)}
    \end{minipage}
  \end{minipage}
  \caption{Warehouse simulation in ROS~2. (a) Gazebo simulation scene.
  (b) Robot paths visualized in RViz.}
  \label{fig:simulation_validation}
\end{figure}

\begin{figure}[!t]
  \centering
  \begin{minipage}{\columnwidth}
    \centering
    \includegraphics[width=0.4784\linewidth]{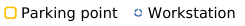}\\[1.6ex]
    \begin{minipage}[t]{0.485\linewidth}\centering
      \includegraphics[width=\linewidth]{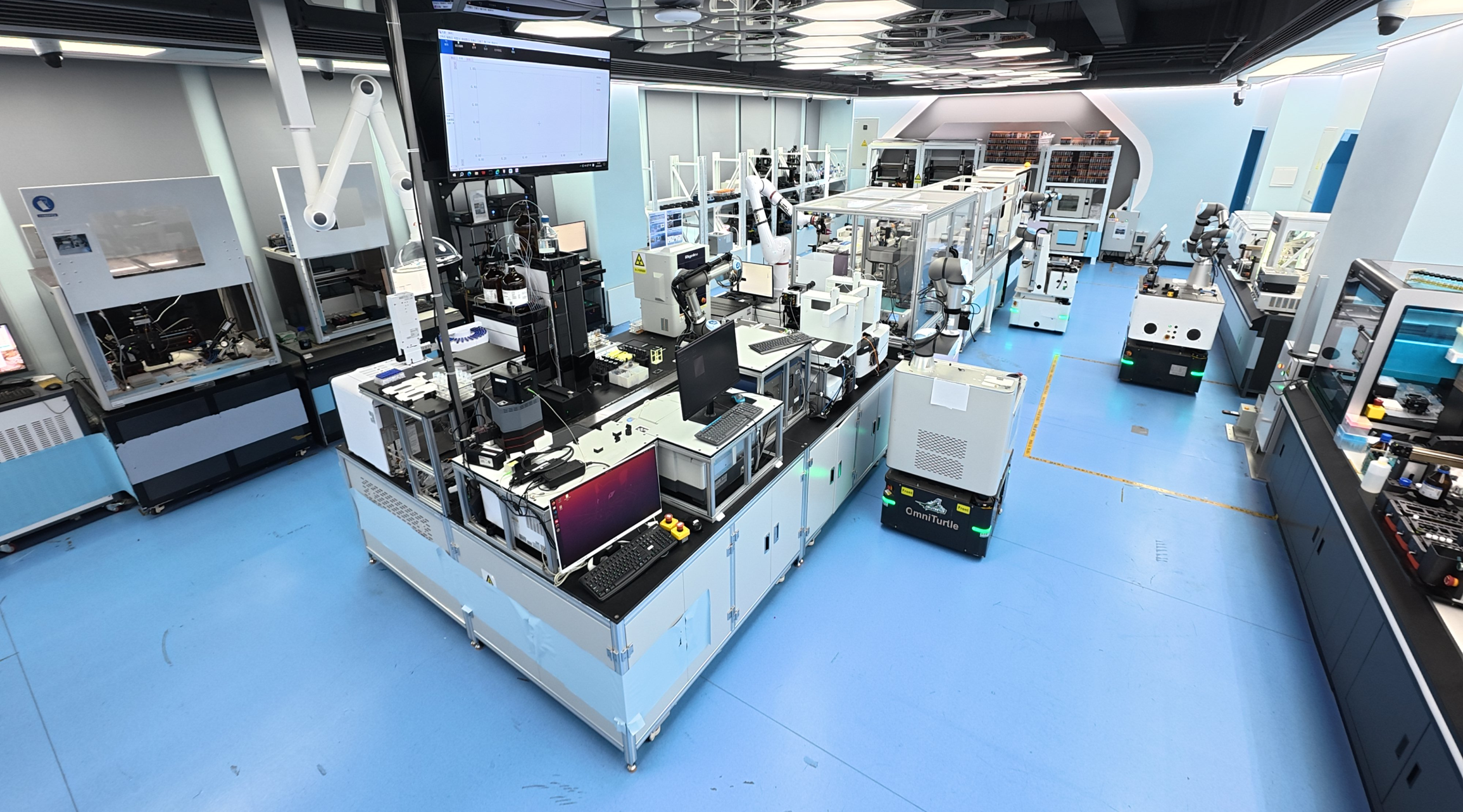}\\[-0.2ex]
      \subfigtext{(a)}
    \end{minipage}\hfill
    \begin{minipage}[t]{0.485\linewidth}\centering
      \includegraphics[width=\linewidth]{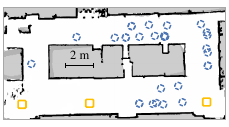}\\[-0.2ex]
      \subfigtext{(b)}
    \end{minipage}
    \par\vspace{0.7ex}
    \begin{minipage}[t]{0.485\linewidth}\centering
      \includegraphics[width=\linewidth]{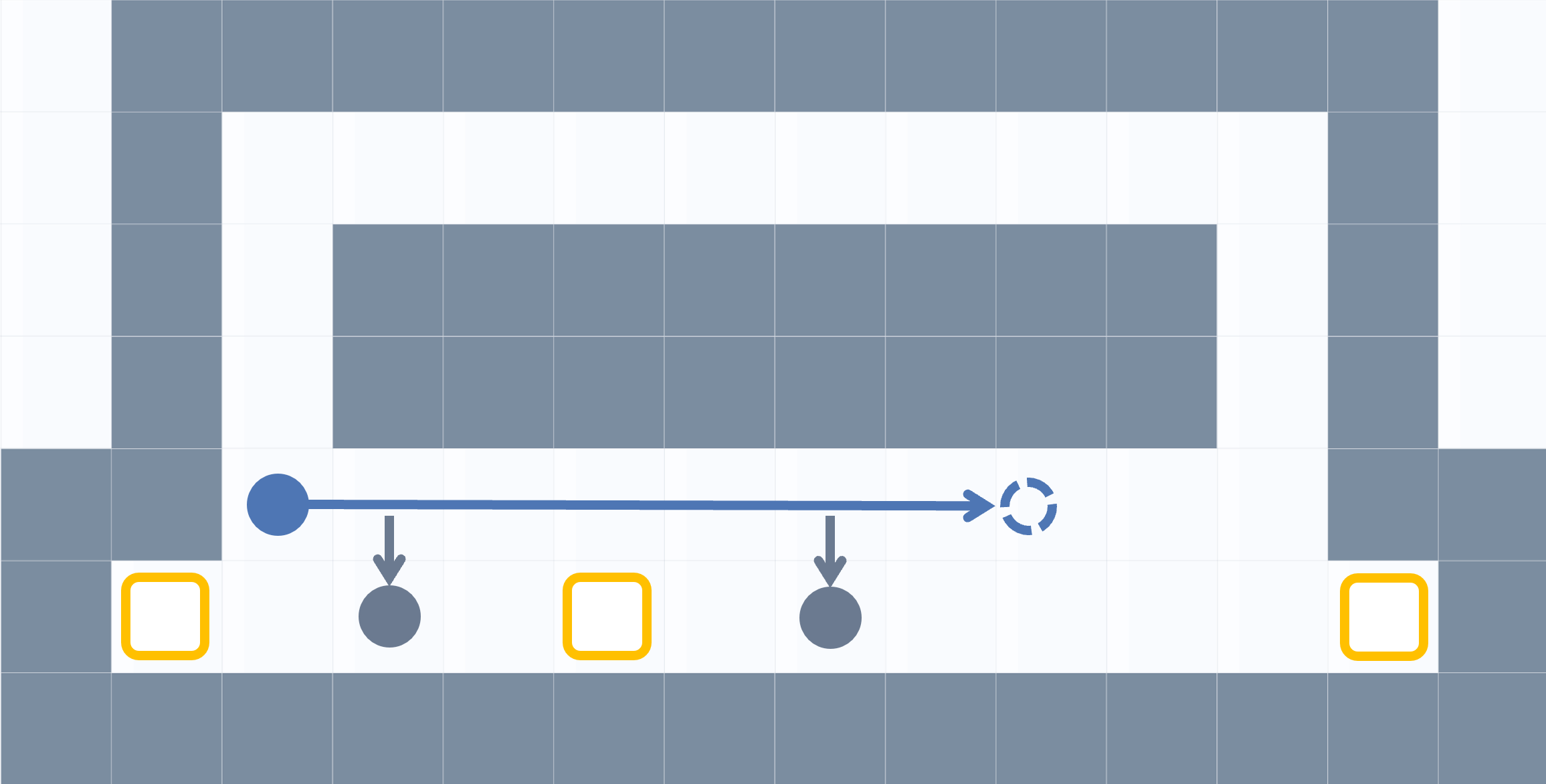}\\[-0.2ex]
      \subfigtext{(c)}
    \end{minipage}\hfill
    \begin{minipage}[t]{0.485\linewidth}\centering
      \includegraphics[width=\linewidth]{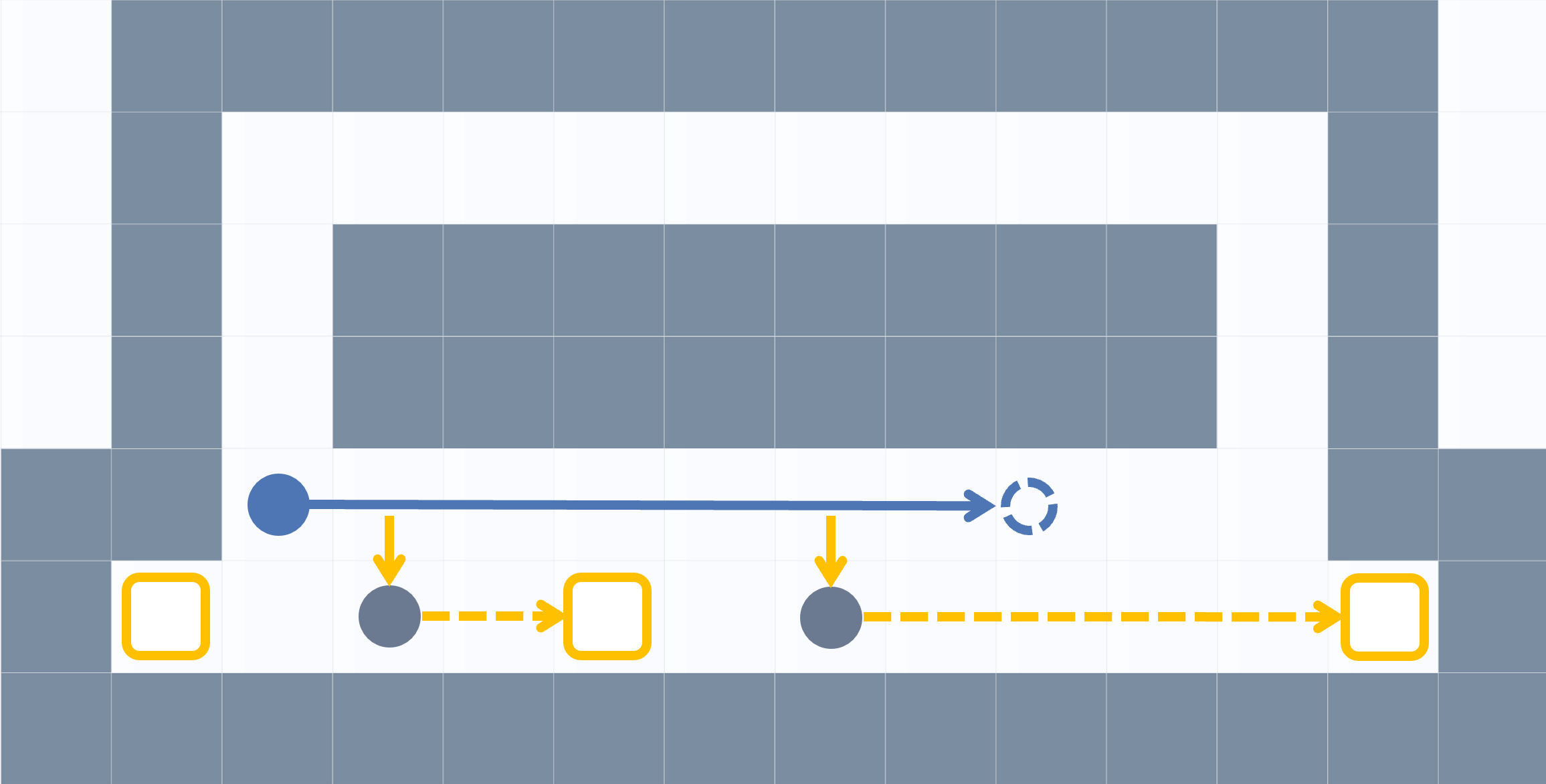}\\[-0.2ex]
      \subfigtext{(d)}
    \end{minipage}
    \par\vspace{0.7ex}
    \begin{minipage}[t]{0.485\linewidth}\centering
      \includegraphics[width=\linewidth]{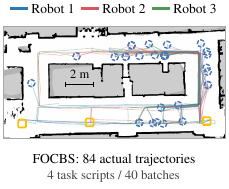}\\[-0.2ex]
      \subfigtext{(e)}
    \end{minipage}\hfill
    \begin{minipage}[t]{0.485\linewidth}\centering
      \includegraphics[width=\linewidth]{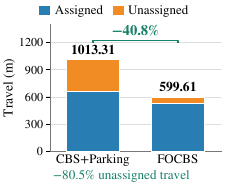}\\[-0.2ex]
      \subfigtext{(f)}
    \end{minipage}
  \end{minipage}
  \caption{Real-robot validation. (a) Laboratory platform. (b) ROS map with workstation and parking targets. (c), (d) Schematics of FOCBS yielding and CBS+Parking, respectively, with blue disks for assigned robots and gray disks for unassigned robots. (e) Recorded FOCBS trajectories across all batches. (f) Measured fleet travel per method, split by per-batch assignment status. Dashed blue circles and yellow squares denote workstation and parking targets.}
  \label{fig:real_robot_validation}
\end{figure}

\subsubsection{Warehouse Simulation}
\label{sec:verification_sim}

Fig.~\ref{fig:simulation_validation} shows the ROS~2/Gazebo simulation on a
20$\times$11 grid with 18 mobile robots and 40 workstations; planning uses
an Intel Core i9-14900K CPU.
The 10 task sequences (115 batches, 1,000 navigation tasks) cover five assigned-robot load levels, with approximately
25--90\% of robots assigned navigation tasks per batch and the remainder
being unassigned robots.
All task sequences were completed, demonstrating coordinated execution of both robot types.
The mean and maximum runtimes per planning call are
\mbox{0.0031 and 0.029~s, respectively}.

\subsubsection{Physical-Robot Experiment}
\label{sec:verification_real}

Fig.~\ref{fig:real_robot_validation}(a) and (b) show the three-robot chemical
laboratory platform and occupancy map, respectively. Planning runs on an
Intel Core i7-13650HX CPU. The ROS~1 robots communicate with the FOCBS
server via MQTT, using FOCBS paths as preferred global references and
eband for local planning. The evaluation covers mobile-base navigation
and excludes chemical experiments and reagent handling by robotic arms.

On this platform, leaving unassigned robots in place as static obstacles
blocks passages between workstations and causes tasks to fail.
FOCBS allows these robots to yield as needed and stop after clearing the
assigned robots' paths (Fig.~\ref{fig:real_robot_validation}(c)).
CBS+Parking sends them to predefined parking areas, using the robots'
charging stations in this experiment (Fig.~\ref{fig:real_robot_validation}(d)).
Parking areas do not obstruct traffic and are typically farther from
the workstations. Both methods complete the same four task sequences, each comprising
10 consecutive batches, with 65 navigation tasks between workstations
in total. One to three robots receive tasks per batch. Comparisons are made over
complete sequences, and travel is grouped by each robot's assignment status
in the corresponding batch.

FOCBS completes all 40 batches without observed inter-robot collisions,
with mean and maximum planning times of 0.067 and 1.365~s, respectively.
Fig.~\ref{fig:real_robot_validation}(e) shows the 84 recorded movement
trajectories, each covering one robot's movement within one batch.

Fig.~\ref{fig:real_robot_validation}(f) shows a 40.8\% reduction in fleet
travel, with savings in every sequence. Most savings come from unassigned
robots, whose total travel decreases by 80.5\%: they wait when they do not
obstruct task execution and, when yielding is required, stop after clearing
the path without continuing to a parking area. Assigned-robot travel also
decreases by 20.2\%. Because each batch inherits the preceding terminal
positions, a robot that remains near the workstations after yielding
travels less on its next assigned task than one departing from a parking
area. The mean batch completion time, measured until all assigned robots
reach their goals, also decreases by 14.7\%.

\section{Conclusion}
\label{sec:conclusion}

This article studied MAPF-UA, which arises when robots with and without
assigned tasks share a workspace. We established an asymmetric
decomposition and proposed FOCBS, which searches over assigned agents with
CBS, routes unassigned agents with a min-cost flow oracle, and converts routing
failures into assigned-agent constraints through flow-conflict branching.
For feasible instances on finite graphs, FOCBS terminates with a sequentially
optimal solution. Benchmark results show that FOCBS scales well with the number of
unassigned agents at fixed assigned-agent load, and ablation
studies confirm the benefits of flow-conflict branching, BZ-SSP, and
start-avoiding tie-breaking. Laboratory experiments show that yielding as
needed reduces fleet travel by 40.8\% relative to parking-based coordination.

Several directions merit further study. FOCBS minimizes $\smu$ only for the
selected $\sst$-optimal assigned-agent paths. Efficiently exploring alternative
$\sst$-optimal A-path sets may further reduce $\smu$.
Combinatorial search over assigned agents limits scalability as tasks
increase, and repeated diagnosis and TEN recomputation dominate hard instances
on bottleneck maps; bounded-suboptimal search and incremental TEN updates may
reduce these costs. Finally, the episode-level objective ignores how terminal
locations affect future tasks, which motivates terminal-aware planning in
lifelong settings.

\bibliographystyle{Bibliography/IEEEtran}
\bibliography{Bibliography/IEEEabrv,references}

\end{document}